\documentclass[10pt,twocolumn,letterpaper]{article}

\usepackage{cvpr}              % To produce the CAMERA-READY version
\definecolor{cvprblue}{rgb}{0.21,0.49,0.74}
\usepackage[pagebackref,breaklinks,colorlinks,allcolors=cvprblue]{hyperref}
\usepackage{multirow}
\usepackage{mathtools} % 提供 \mathclap 命令
\usepackage{bbding}
\usepackage{algorithmic}
\usepackage{algorithm}
\def\paperID{} % *** Enter the Paper ID here
\def\confName{CVPR}
\def\confYear{2026}

\title{Bridging Day and Night: Unsupervised Cross-Domain Re-Identification with Synergistic Prompt and Prototype Learning}

\author{
Jiyang Xu$^{1*}$ \quad
Rui Liu$^{1*}$ \quad
Hang Dai$^{1\dagger}$\\[0.6em]
$^{1}$School of Computer Science, Wuhan University, Wuhan, China\\
{\tt\small
jiyangxu@whu.edu.cn, liurui0643@whu.edu.cn, daihang@whu.edu.cn
}
}

\begin{document}
\maketitle
\begingroup
\renewcommand{\thefootnote}{}
\footnote{%
\begin{tabular}{@{}ll@{}}
$^{*}$Equal contribution, $^{\dagger}$Corresponding author \\
\url{https://github.com/jyxucs/USL-DN-ReID}
\end{tabular}
}
\addtocounter{footnote}{-1}
\endgroup

\begin{abstract}
Cross-domain day–night re-identification (ReID) is fundamentally challenged by the substantial visual appearance discrepancies between daytime and nighttime scenes. Existing fully supervised methods rely heavily on labor-intensive annotations, which are costly and exhibit limited generalization across domains.  
In this work, we investigate unsupervised day–night ReID and propose a novel framework that synergistically combines prompt learning and prototype-based representation learning to associate identities across domains without requiring manual labels. Our approach follows a progressive two-stage training strategy. In the first stage, we exploit the vision–language model to generate instance-specific textual prompts in an annotation-free manner. We employ an instance-level alignment mechanism to embed visual features and textual prompts into a unified semantic space, aligning unlabeled day/night images with learnable prompts via instance-aware dynamic-bias adaptation. In the second stage, we construct domain-specific prototype memory banks and introduce two complementary modules: i) an intra-domain identity association module to enhance feature discriminability within each domain, and ii) a cross-domain prototype matching module to reliably identify positive and negative prototype pairs, thereby establishing robust identity correspondences across day and night.  
Extensive experiments on public benchmarks validate the effectiveness of our method. Under the unsupervised setting, our framework attains Rank-1 accuracy comparable to state-of-the-art fully supervised methods.%, demonstrating its potential for practical deployment.
\end{abstract}
    
\section{Introduction}
\label{sec:intro}

Day–night vehicle re-identification (DN-ReID)~\cite{ref_1,ref_2, ref_25} constitutes a pivotal component of intelligent surveillance systems and has recently attracted substantial research attention. Despite notable progress, DN-ReID remains intrinsically challenging due to the pronounced domain shift induced by drastic illumination changes between day and night, and inherent viewpoint variations. While fully supervised approaches~\cite{ref_2,ref_4,ref_5, ref_49} have established strong performance benchmarks, their reliance on extensive manual annotations limits scalability and practical deployment, motivating the investigation of DN-ReID under an unsupervised learning paradigm, which remains largely underexplored.

Unsupervised day–night vehicle re-identification (USL-DN-ReID) encounters substantially greater challenges than its supervised counterpart. The primary difficulties stem from: i) the lack of reliable supervisory signals, ii) the pronounced domain discrepancies between day and night imagery, which induce significant feature distribution shifts, and iii) the inherent complexity of establishing accurate cross-domain identity associations. Recent advances in vision–language models (VLM)~\cite{ref_6,ref_8,ref_48} provide promising avenues to mitigate these challenges. Models such as CLIP~\cite{ref_8} demonstrate strong cross-modal semantic alignment capabilities, enabling visual and textual features to be projected into a shared embedding space. CLIP-ReID~\cite{ref_9} employs a prompt-learning mechanism~\cite{ref_10} to generate structured textual descriptions for each identity using templates of the form ``A photo of a $[V_1][V_2]\ldots[V_M]$ vehicle”, providing consistent textual cues for visual samples and facilitating semantic alignment across day–night domains. However, these methods~\cite{ref_9,ref_11,ref_12} fundamentally depend on ground-truth identity labels for prompt construction, and their static prompt representations fail to adapt to dynamically evolving pseudo-label distributions in unsupervised settings, rendering them inapplicable to USL-DN-ReID.

Recent progress in unsupervised visible–infrared re-identification (USL-VI-ReID)~\cite{ref_14,ref_15,ref_45,ref_46} demonstrates the feasibility of learning cross-modal associations without manual identity annotations. These methods commonly leverage clustering-based pseudo-label generation combined with cross-modal contrastive learning to align heterogeneous features. Representative approaches such as SDCL~\cite{ref_15} and MMM~\cite{ref_16} primarily attribute the modality gap to feature distribution shifts induced by spectral differences, and aim to mitigate it through feature-level alignment and shared prototype learning. Directly transferring these strategies to USL-DN-ReID is non-trivial. Unlike VI-ReID, where modality discrepancies primarily arise from spectral differences, DN-ReID is affected by complex environmental illumination, with nighttime imagery frequently impacted by strong light interference and low-light noise, resulting in more severe and heterogeneous domain shifts~\cite{ref_2}. Most existing methods circumvent explicit instance-to-instance association across domains, instead adopting proxy-based prototype learning~\cite{ref_13,ref_16,ref_17,ref_18,ref_43} to establish cross-modal connections. This motivates a reformulation of the problem: rather than focusing solely on instance-level correspondences, it is essential to model relationships among cross-domain prototypes to capture the intrinsic structure of identity representations across varying illumination conditions.

Building upon these insights, we propose a novel USL-DN-ReID framework that synergistically integrates instance-aware prompt learning with cross-domain prototype consistency modeling to achieve robust day–night association. Our approach capitalizes on the strong semantic alignment capabilities of VLM while explicitly addressing their limitations in unsupervised scenarios. In Stage-1, we introduce a dynamic-bias network that adaptively aligns textual representations with individual visual instances, thereby establishing reliable image–text correspondences without identity annotations. In Stage-2, domain-specific prototype memory banks are constructed for daytime and nighttime domains, and a cross-domain prototype matching strategy is employed to identify positive and negative prototype pairs across domains. Our contributions are:
\begin{enumerate}[i)]
\item We propose a synergistic learning framework for USL-DN-ReID that effectively integrates prompt-based semantic alignment with prototype-based representation learning.
\item We design an instance-aware prompt learning module tailored for unsupervised settings, which dynamically generates discriminative textual prompts for unlabeled vehicle images.
\item We introduce a cross-domain prototype matching mechanism with bidirectional matching, enabling robust and reliable cross-domain identity correspondences in the absence of annotations.
\item Extensive experiments on public benchmarks demonstrate the superiority of our USL-DN-ReID framework, achieving competitive performance relative to state-of-the-art fully supervised approaches.
\end{enumerate}

% Furthermore, samples captured under day and night conditions exhibit many-to-many dynamic correspondences rather than the relatively stable one-to-one mappings typical in VI-ReID. 
\section{Related Work}

\textbf{Prompt Learning for ReID.}  
Recent advances in vision–language pre-trained models\cite{ref_8,ref_20,ref_40}, such as CLIP~\cite{ref_8}, have demonstrated strong zero-shot transfer capabilities in downstream tasks. The effectiveness of these models critically depends on high-quality textual descriptions for image–text alignment. To mitigate the reliance on manual annotations, Zhou et al.~\cite{ref_10} introduced prompt learning, wherein learnable vectors are optimized to semantically align textual embeddings with visual features.
Building upon this paradigm, prompt learning \cite{ref_19,ref_44,ref_36,ref_45} has been increasingly applied to person and vehicle re-identification. Li et al.~\cite{ref_9} pioneered the integration of pre-trained CLIP with textual prompts describing visual identities, employing a two-stage training strategy to enhance cross-modal alignment. Subsequent works~\cite{ref_11,ref_21,ref_22,ref_23} further extended this approach by designing automated text templates to enrich visual feature expressiveness. For instance, CSDN~\cite{ref_11} adapts prompt learning for VI-ReID, enabling high-level semantic feature matching across modalities, while Zhai et al.~\cite{ref_22} systematically investigate the effects of explicit versus implicit textual representations on ReID performance.
Despite these advances, existing prompt-based ReID methods largely assume the availability of identity labels or paired data for prompt optimization, which constrains their applicability in unsupervised settings. Moreover, static prompt designs lack the flexibility to adapt to evolving cross-domain feature distributions. These limitations motivate our exploration of adaptive, instance-aware prompt learning tailored for unsupervised day–night vehicle re-identification.

\begin{figure*}[!h]
    \centering
    \includegraphics[width=\textwidth]{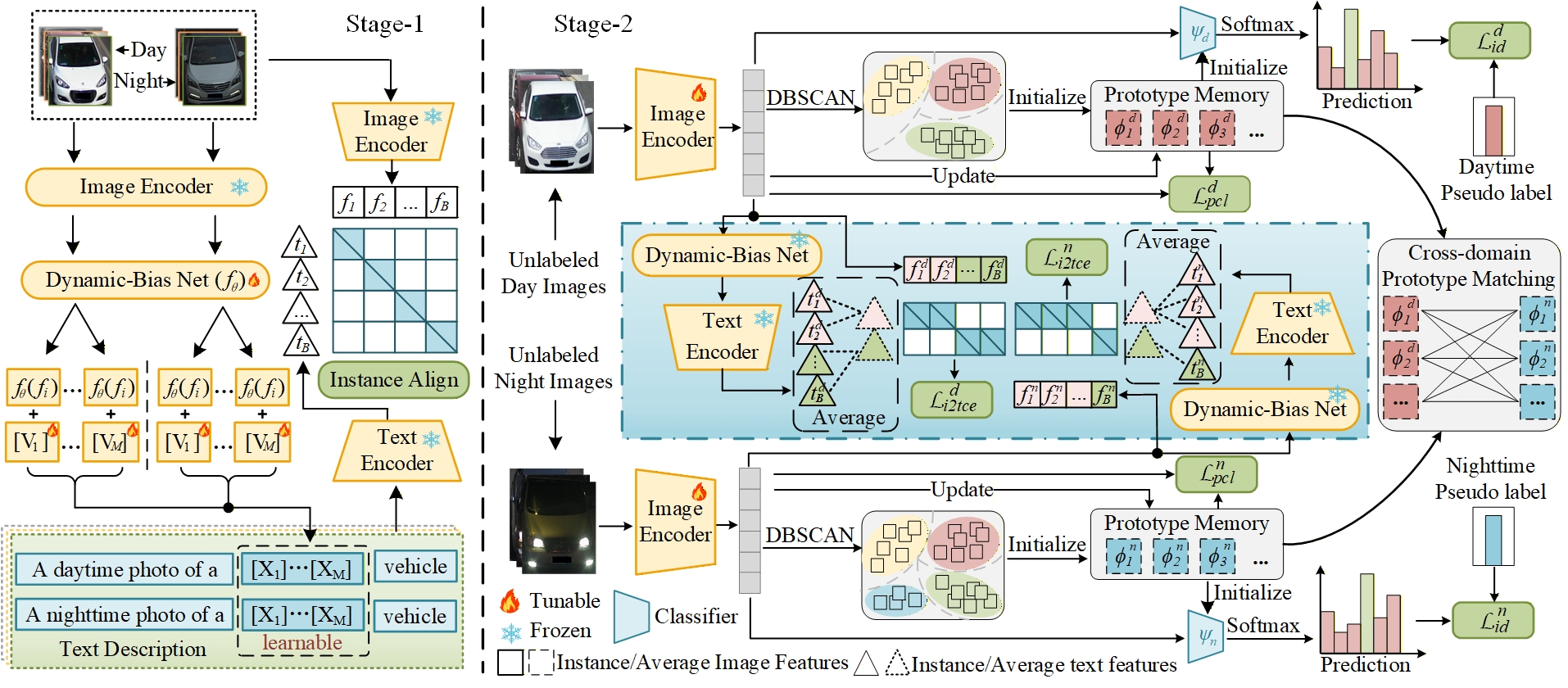}
     \caption{Overview of the proposed synergistic prompt and prototype learning framework. The framework is trained in two stages: \textbf{Stage-1} freezes both the image and text encoders, and aligns unlabeled day/night images with learnable textual prompts via instance-aware dynamic-bias adaptation. \textbf{Stage-2} activates the image encoder and incorporates: (i) intra-domain identity association using prototype memory banks to enhance feature discriminability within each domain, and (ii) cross-domain prototype matching to identify reliable positive and negative prototype pairs, establishing annotation-free identity correspondences across day and night domains.}

    \label{img_1}
\end{figure*}

\noindent\textbf{Unsupervised Visible–Infrared ReID.}  
Unsupervised visible–infrared re-identification (USL-VI-ReID)~\cite{ref_13,ref_24, ref_41} has emerged as a practical alternative to fully supervised methods, aiming to learn modality-invariant representations without relying on manual identity annotations. Early approaches predominantly adopt two-stage training paradigms based on clustering-generated pseudo-labels. For instance, H2H~\cite{ref_24} and CHCR~\cite{ref_18} first learn discriminative features within each modality and subsequently extract shared semantic knowledge across modalities to enhance cross-modal alignment.
With the widespread integration of contrastive learning frameworks such as MoCo~\cite{ref_26} and prototype-based representation learning~\cite{ref_27}, recent methods have increasingly embraced graph- and prototype-centric strategies. PGM~\cite{ref_28} introduces a progressive graph matching framework that iteratively refines identity correspondences by alternating between cross-modal graph association and feature updating. Similarly, Shi et al.~\cite{ref_29} construct multiple prototype representations and leverage progressive contrastive learning to capture more discriminative and diverse feature patterns across modalities.
Despite these advancements, the underlying differences in imaging characteristics and environmental variability between infrared imagery and day–night scenes preclude the direct transfer of USL-VI-ReID techniques to the USL-DN-ReID task. These discrepancies necessitate the development of methods tailored to address the complex illumination-induced domain gaps inherent to USL-DN-ReID.

\section{Proposed Method}

\textbf{Overview.}
The proposed USL-DN-ReID framework operates through two progressive training stages, as depicted in Figure~\ref{img_1}. Given an unlabeled day–night image dataset 
$\mathcal{I} = \{\{\mathcal{I}^d_i\}_{i=1}^{N_d} \cup \{\mathcal{I}^n_i\}_{i=1}^{N_n}\}$, where $N_d$ and $N_n$ denote the numbers of unlabeled images in the daytime and nighttime domains, respectively, the images are first processed by a vision encoder with frozen parameters to extract instance-level visual features.

\textbf{Stage-1} establishes an initial semantic alignment with a frozen CLIP model. The proposed instance-aware prompt learning module, empowered by a dynamic-bias network and optimized via a bidirectional contrastive loss, generates adaptive textual descriptions for each unlabeled day–night image. This enables the projection of visual and textual representations into a shared latent space, facilitating robust image–text correspondence without identity annotations.

\textbf{Stage-2} focuses on prototype-driven identity association and consists of two components: the Intra-domain Identity Association (IIA) module and the Cross-domain Prototype Matching Learning (CPML) module. The IIA module constructs domain-specific prototype memory banks by generating pseudo-labels through clustering, thereby enhancing feature discriminability within each domain. The CPML module identifies reliable cross-domain correspondences by mining and aligning positive and negative prototype pairs between daytime and nighttime prototype banks, effectively bridging the domain gap and promoting robust identity association across illumination conditions.

% The second stage comprises two key components: an Intra-domain Identity Association (IIA) module and a Cross-Prototype Matching Learning (CPML) module. Pseudo-labels ${y^d}$ and ${y^n}$ are first generated for samples from the day and night domains, respectively, using a clustering algorithm. These pseudo-labels are then used to initialize the corresponding cluster memory banks. The IIA module leverages these memory banks to model intra-domain feature distributions, while incorporating the text prompts learned in the first stage to enhance feature discriminability. Meanwhile, the CPML module constructs cross-domain positive and negative sample pairs through a memory-based matching strategy, establishing reliable identity correspondences that promote semantic consistency and facilitate knowledge transfer between the day and night domains.

\subsection{Instance-aware Prompt Learning}
Leveraging large-scale vision–language models in cross-modal alignment, recent works such as CLIP-ReID~\cite{ref_9} have employed prompt learning strategies (e.g., CoOp~\cite{ref_10}) to generate annotation-free textual descriptions that enhance visual representation learning. Despite their effectiveness, these approaches are not directly applicable to USL-DN-ReID for two key reasons: i) without identity annotations, stage-one training cannot assign discriminative textual descriptions to individual instances; ii) global prompt vectors alone are insufficient to capture fine-grained, instance-specific cues within the same identity cluster.

As shown in Figure~\ref{img_1}, during stage-1 training, we construct instance-level textual descriptions $\{T^d, T^n\}$ for each image in $\{\mathcal{I}^d, \mathcal{I}^n\}$. Each description follows the template: ``A daytime/nighttime photo of a $[V_1][V_2]\dots[V_M]$ vehicle", where $V_M$ denotes learnable prompt vectors following the CoOp framework and $M$ is the number of such vectors. In the USL setting, pseudo-labels evolve dynamically during training, whereas CoOp’s prompt vectors remain static, leading to suboptimal representation for stage-2 learning. 

To address this, inspired by CoCoOp~\cite{ref_36}, we introduce an \emph{instance-aware prompt adaptation} strategy. Specifically, we first extract static visual features $f_i$ for each instance using a frozen visual encoder. These features are projected into the prompt semantic space via a Dynamic-bias Net, implemented as a lightweight MLP $f_\theta(\cdot)$, producing instance-specific bias terms that dynamically modulate textual prompt representations:
\begin{equation}
    X_m(f_i) = V_m + f_\theta(f_i), \quad m \in [1, M].
\label{eq_1}
\end{equation}
The resulting text prompts are then encoded with the frozen text encoder to obtain instance-aligned textual features $t_i$, enabling fine-grained, identity-consistent cross-modal alignment.
 
Following the CLIP paradigm, we perform instance-level alignment between visual and textual feature pairs $\langle f_i, t_i\rangle$ within a shared semantic space. The model is optimized via a bidirectional contrastive learning objective, encompassing both image-to-text and text-to-image components, denoted as $\mathcal{L}_{i2t}$ and $\mathcal{L}_{t2i}$, respectively. Formally, the objectives are defined as:
\begin{equation}
    \mathcal{L}_{i2t} = - \log \frac{
    \exp\left( S(f_i, t_i) / \tau \right)
}{
    \sum_{j=1}^B \exp\left( S(f_i, t_j) / \tau \right)
}, 
\label{eq_2}
\end{equation}
\begin{equation}
    \mathcal{L}_{t2i} = - \log \frac{
    \exp\left( S(t_i, f_i) / \tau \right)
}{
    \sum_{j=1}^B \exp\left( S(t_i, f_j) / \tau \right)
},
\label{eq_3}
\end{equation}
where $S(\cdot, \cdot)$ denotes a similarity function between two feature vectors, $B$ represents the batch size, and $\tau$ is a temperature hyperparameter. 
This bidirectional formulation enforces tight instance-level alignment, ensuring that each image feature is closely associated with its corresponding textual prompt while remaining discriminative relative to other instances within the batch.

\subsection{Intra-domain Identity Association}
During stage-2 training, unlabeled images from the daytime and nighttime domains are processed through a learnable image encoder at each epoch. Subsequently, the DBSCAN algorithm is applied to generate dynamic pseudo-labels $\{y^d, y^n\}$ for each instance, which are updated at every epoch according to the latest clustering results. 

Prototype memory banks are then initialized by computing the mean feature representation of instances within each cluster, yielding the prototype features $\{\phi^d, \phi^n\}$:
\begin{equation}
\phi_{y_i^d}^{d} = \frac{1}{|C_{y_i^d}^{d}|} \sum_{f_i^{d}\in C_{y_i^d}^{d}} f_i^{d},
\quad\phi_{y_i^n}^{n} = \frac{1}{|C_{y_i^n}^{n}|} \sum_{f_i^{n}\in C_{y_i^n}^{n}} f_i^{n},
\label{eq_4}
\end{equation}
where $f_i^d$ and $f_i^n$ denote the instance-level features of the daytime and nighttime domains, respectively. $C_{y_i^d}^d$ and $C_{y_i^n}^n$ represent the sets of instances assigned to pseudo-labels $y_i^d$ and $y_i^n$, and $|\cdot|$ denotes the cardinality of each set.

To ensure stability and smooth evolution of prototypes during training, we update the prototype features using a momentum-based strategy:
\begin{equation}
\phi_{y_i^\nabla}^{\nabla, \delta} \leftarrow \alpha \phi_{y_i^\nabla}^{\nabla, \delta-1} + (1-\alpha) f_i^\nabla, \quad \nabla \in \{d,n\},
\label{eq_6}
\end{equation}
where $\alpha \in [0,1)$ is the momentum factor controlling the update rate, and $\delta$ denotes the current training iteration. This formulation ensures that prototype representations evolve smoothly while incorporating newly observed instance features in the proposed IIA.

\textbf{Prototype Contrastive Loss.} 
To enhance alignment between instance-level visual features and their corresponding domain-specific prototypes, while simultaneously promoting discrimination from other prototypes within the same domain, we employ a prototype contrastive loss:
\begin{equation}
{\cal L}_{pcl} = -\sum_{\nabla\in\{d,n\}} \log \frac{\exp(S(f_i^\nabla, \phi_+^\nabla)/\tau)}{\sum_{c=1}^{C^\nabla} \exp(S(f_i^\nabla, \phi_c^\nabla)/\tau)},
\label{eq_7}
\end{equation}
where $\phi_+$ denotes the prototype associated with the query feature $f_i^\nabla$, and $C^\nabla$ is the total number of prototypes in domain $\nabla \in \{d,n\}$.

\textbf{Prototype-driven Classification.} 
In the unsupervised setting with dynamically evolving pseudo-labels, we append domain-specific classifiers $\{\psi_d, \psi_n\}$ to the last layer of the image encoder for daytime and nighttime domains, respectively. 
Before each training epoch, these classifiers are initialized with the corresponding prototypical features stored in the memory banks $\{\phi^d, \phi^n\}$. 
The classifiers project instance features onto centroid-oriented predictions, which are supervised by the cross-entropy loss $H(\cdot)$ with the pseudo-labels:
\begin{equation}
{\cal L}_{id}^d = \frac{1}{N_d}\sum_{i=1}^{N_d} H(\mathrm{softmax}(\psi_d(f_i^d)), y_i^d),
\label{eq_8}
\end{equation}
\begin{equation}
{\cal L}_{id}^n = \frac{1}{N_n}\sum_{i=1}^{N_n} H(\mathrm{softmax}(\psi_n(f_i^n)), y_i^n),
\label{eq_9}
\end{equation}
where $N_d$ and $N_n$ denote the number of instances in the daytime and nighttime domains, respectively.

\textbf{Image–Text Alignment.} 
Textual descriptions obtained from stage-1 are directly reused in stage-2. While the learned prompt vectors and the stage-1 dynamic-bias network remain frozen, the instance-specific bias terms are updated dynamically to reflect evolving visual features during stage-2. 
We adopt the CLIP-ReID~\cite{ref_9} image-to-text cross-entropy loss $\mathcal{L}_{i2tce}$ for optimization. 
Under USL setting, computing the same-class textual centroids for each query $f_i$ in every iteration is computationally expensive. To address this, we compute $\mathcal{L}_{i2tce}$ in a batch-wise manner:
\begin{equation}
{\cal L}_{i2tce}^d = -\sum_{i=1}^B q_i^d \cdot \log \frac{\exp(S(f_i^d, \Phi(t_i^d,y_i^d)))}{\sum_{j=1}^{\hat{C}^d} \exp(S(f_i^d, \Phi(t_j^d,y_j^d)))},
\label{eq_10}
\end{equation}
where $\hat{C}^d$ denotes the set of textual centroids within a batch for the daytime domain, $q_i^d$ represents the smoothed pseudo-label of $y_i^d$ in the batch, and $\Phi(t_i^d, y_i^d)$ denotes the centroid of text features $t_i^d$ corresponding to pseudo-label $y_i^d$, computed analogously to Eq.~\ref{eq_4}. The nighttime domain loss $\mathcal{L}_{i2tce}^n$ is computed in the same manner.

\begin{figure}[!h]
    \centering
    \includegraphics[width=.5\textwidth]{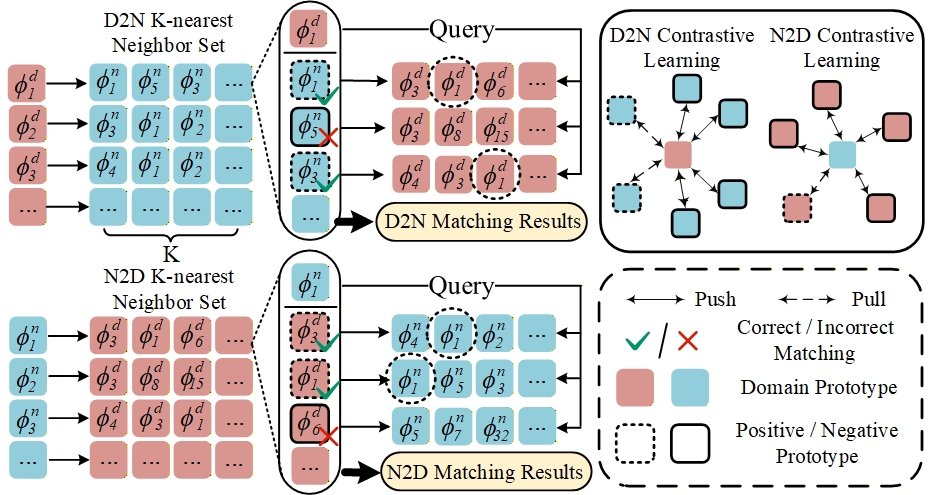}
    \caption{Illustration of the Cross-Domain Prototype Matching.}
    \label{img_2}
\end{figure}
\subsection{Cross-Domain Prototype Matching}
While the Intra-domain Identity Association (IIA) module focuses on modeling intra-domain identity relationships by optimizing feature distributions within each domain, it does not explicitly establish correspondences across domains. To enable robust cross-domain associations between daytime and nighttime identities, we introduce the Cross-Prototype Matching Learning (CPML), as illustrated in Figure~\ref{img_2}.

\textbf{Cross-Domain Prototype Dictionary Construction.}
We construct a cross-domain prototype dictionary by leveraging the domain-specific prototype sets $\phi^d = \{\phi^d_1, \dots, \phi^d_{C^d}\}$ for daytime and $\phi^n = \{\phi^n_1, \dots, \phi^n_{C^n}\}$ for nighttime. Specifically, we compute a similarity matrix of size $C^d \times C^n$ between all cross-domain prototypes. For each prototype, we select the top-$k$ most similar cross-domain prototypes to form its k-nearest neighbor set ${\cal R}(\cdot)$:
\begin{equation}
{\cal R}(\phi^d_i) = \mathrm{Top}\text{-}k\big(S(\phi^d_i, \{\phi^n_j\}_{j=1}^{C^n})\big),
\label{eq_2}
\end{equation}
\begin{equation}
{\cal R}(\phi^n_j) = \mathrm{Top}\text{-}k\big(S(\phi^n_j, \{\phi^d_i\}_{i=1}^{C^d})\big),
\label{eq_3}
\end{equation}
where $S(\cdot,\cdot)$ denotes a similarity measure between two prototype features.

\textbf{Prototype Matching.} 
The constructed cross-domain dictionary provides a reference for establishing reliable identity correspondences. 
Given a daytime prototype $\phi^d_i$ and a candidate nighttime prototype $\phi^n_j \in {\cal R}(\phi^d_i)$, we consider $\langle \phi^d_i, \phi^n_j \rangle$ a positive pair if $\phi^d_i \in {\cal R}(\phi^n_j)$, indicating mutual nearest neighbors. Otherwise, $\phi^n_j$ is treated as a hard negative of $\phi^d_i$. 
The k-nearest neighbor set of each prototype is partitioned into positive $\mathcal{P}(\cdot)$ and negative $\mathcal{N}(\cdot)$ subsets:
\begin{equation}
\langle \phi^d_i, \{\phi^n_j\}_{j=1}^{k} \rangle = 
\begin{cases}
{\cal P}(\phi^d_i) = \{\phi^n_j \mid \phi^d_i \in {\cal R}(\phi^n_j)\},\\
{\cal N}(\phi^d_i) = \{\phi^n_j \mid \phi^d_i \notin {\cal R}(\phi^n_j)\},
\end{cases}
\label{EQ_1}
\end{equation}
where $\mathcal{P}(\phi^d_i) \cup \mathcal{N}(\phi^d_i) = k$. The same partitioning is applied for nighttime prototypes $\phi^n_j$ and their neighbor sets ${\cal R}(\phi^n_j)$.

\textbf{Cross-Domain Prototype Matching Loss.} 
The sets of mutually matched positive prototypes across domains are symmetric, i.e., $\sum \mathcal{P}(\phi^d) = \sum \mathcal{P}(\phi^n)$. We denote unmatched daytime and nighttime prototypes as negative sets $\mathcal{N}^{d2n}$ and $\mathcal{N}^{n2d}$, respectively. 
Using these positive and negative sets, we define the cross-domain prototype matching loss based on InfoNCE:
\begin{align}
\small
\mathcal{L}_{cpml} &= 
-\frac{1}{2|\mathcal{P}|} \sum_{(i,j)\in \mathcal{P}} \Biggl( \log \frac{\exp\Bigl(S(\phi^d_i, \phi^n_j)/\tau\Bigr)}
{\sum\limits_{(i,j^-) \in \mathcal{N}^{d2n} \cup \mathcal{P}} \exp\Bigl(S(\phi^d_i, \phi^n_{j^-})/{\tau}\Bigr)} \notag \\
&\quad + \log \frac{\exp\Bigl(S(\phi^n_i, \phi^d_j)/\tau\Bigr)}
{\sum\limits_{(i,j^-) \in \mathcal{N}^{n2d} \cup \mathcal{P}} \exp\Bigl(S(\phi^n_i, \phi^d_{j^-})/{\tau}\Bigr)} \Biggr)
\end{align}
% \begin{align}
% \small
% &\mathcal{L}_{cpml} = 
% \log \frac{\exp\Bigl(S(\phi^n_i, \phi^d_j)/\tau\Bigr)}
% {\sum\limits_{(i,j^-) \in \mathcal{N}^{n2d}} \exp\Bigl(S(\phi^n_i, \phi^d_{j^-})/{\tau}\Bigr)} \\
% &\quad -\frac{1}{2|\mathcal{P}|} \sum_{(i,j)\in \mathcal{P}} \Biggl( \log \frac{\exp\Bigl(S(\phi^d_i, \phi^n_j)/\tau\Bigr)}
% {\sum\limits_{(i,j^-) \in \mathcal{N}^{d2n}} \exp\Bigl(S(\phi^d_i, \phi^n_{j^-})/{\tau}\Bigr)} \notag \Biggr)
% \end{align}

\subsection{Loss Objective}

\textbf{Training.} The proposed framework is optimized in a progressive two-stage manner.

\noindent \textbf{Stage-1.} Both the visual and text encoders remain frozen. The model is trained using a bidirectional image-to-text contrastive objective to align instance-level visual and textual features. The stage-1 training objective is defined as:
\begin{equation}
{\cal L}_{\mathrm{stage1}} = \sum_{\nabla \in \{d,n\}} \left( {\cal L}_{i2t}^{\nabla} + {\cal L}_{t2i}^{\nabla} \right),
\label{eq:stage1_loss}
\end{equation}
where ${\cal L}_{i2t}$ and ${\cal L}_{t2i}$ are the image-to-text and text-to-image contrastive losses, respectively, for each domain $\nabla$.

\begin{table*}[!h]
	\belowrulesep=0pt
	\aboverulesep=0pt
	\centering
	\resizebox{\linewidth}{!}{
		\begin{tabular}{c|c|ccc|ccc|ccc|ccc}
			\toprule[2pt] %change the first line to \toprule
			\multicolumn{2}{c|}{\multirow{3}{*}{\parbox[c]{2cm}{\centering \textbf{Methods}}}} & \multicolumn{6}{c|}{\textbf{DN-348}} & \multicolumn{6}{c}{\textbf{DN-Wild}}\\
			\cline{3-14} 
			\multicolumn{1}{c@{}}{}& &\multicolumn{3}{c|}{Day-to-Night} & \multicolumn{3}{c|}{Night-to-Day}& \multicolumn{3}{c|}{Day-to-Night} & \multicolumn{3}{c}{Night-to-Day}  \\
			\cline{3-14}
			\multicolumn{1}{c@{}}{}& & Rank1 & Rank5 & mAP & Rank1 & Rank5 & mAP & Rank1 & Rank5 & mAP & Rank1 & Rank5 & mAP \\
			% \toprule[2pt] %change the second line to midrule
            \hline
		     \multirow{5}{*}{\rotatebox{90}{FSL}} 
                &CAJ~\cite{ref_30}(ICCV'21) &0.660 & 0.819 & 0.464 & 0.739 & 0.884 & 0.453 & 0.499 & 0.981 & 0.392 & 0.487 & 0.955 & 0.385 \\
                % &TransReID&ICCV'21&Vit&82.0& 97.1& - & - & 85.2&97.5 & - &-&-&-&-&-\\
                &AGW~\cite{ref_31}(TPAMI'22) &0.681 & 0.821 & 0.465 & 0.731 & 0.894 & 0.443 & 0.483 & 0.960 & 0.387 & 0.491 & 0.963 & 0.386\\
                &DCL~\cite{ref_5}(ACMMM'22) & 0.675 & 0.813 & 0.443 & 0.789 & 0.920 & 0.428 & 0.499 & 0.979 & 0.348 & 0.472 & 0.943 & 0.342\\
                &PMT~\cite{ref_4}(AAAI'23) & 0.663 & 0.820 & 0.470 & 0.760 & 0.907 & 0.461 & 0.491 & 0.955 & 0.327 & 0.444 & 0.902 & 0.337\\
                &DNDM~\cite{ref_2}(CVPR'24)& 0.707 & 0.842 & 0.475 & 0.803 & 0.926 & 0.462 & 0.512 & 0.987 & 0.405 & 0.495 & 0.955 & 0.400\\

			\hline
                \multirow{7}{*}{\rotatebox{90}{USL}}
                
                & PGM~\cite{ref_28}(CVPR'23) &0.573&0.742&0.315&0.657&0.814&0.328&0.412&0.868&0.192&0.397&0.818&0.223\\
                & RPNR~\cite{ref_32}(ACMMM'24) &0.587&0.752&0.347&0.710&0.872&0.353&\underline{0.467}&\underline{0.925}&\underline{0.248}&\underline{0.442}&\underline{0.898}&\underline{0.261}\\
                & PCLHD~\cite{ref_29}(NeurIPS'24) &\underline{0.648}&\underline{0.787}&0.387&\underline{0.784}&\textbf{0.909}&\underline{0.395}&0.453&0.913&0.232&0.435&0.886&0.251 \\
                & TokenMatcher~\cite{ref_33}(AAAI'25) & 0.549&0.729&0.323&0.702&0.877&0.331&0.455&0.904&0.218&0.415&0.856&0.229\\
                & NULC~\cite{ref_34}(AAAI'25) &0.630&0.777&\underline{0.389}&0.753&0.896&0.384&0.455&0.916&0.229&0.417&0.863&0.237\\
                & PCA~\cite{ref_35}(TIFS'25) &0.633 &0.780 & 0.383 & 0.746 &\underline{0.901} &0.389&0.442&0.903&0.207&0.426&0.870&0.247\\
		    \cline{2-14} 
                  & Ours  & \textbf{0.707} & \textbf{0.823} & \textbf{0.423} & \textbf{0.796} & 0.898 & \textbf{0.437} & \textbf{0.499} & \textbf{0.958} & \textbf{0.282} & \textbf{0.476} & \textbf{0.930} & \textbf{0.295}\\

			\bottomrule[2pt] %change the third line to bottomrule

	\end{tabular}}
	\caption{Performance comparison with state-of-the-art methods on supervised and unsupervised tasks. FSL and USL denote fully-supervised and unsupervised learning methods, respectively. \textbf{Bold} and \underline{underline} indicate the best and second-best results.}
    \label{table_1}
\end{table*}

\begin{table*}[!h]
	\belowrulesep=0pt
	\aboverulesep=0pt
	\centering
	\resizebox{\linewidth}{!}{
		\begin{tabular}{c|ccc|ccc|ccc|ccc}
			\toprule[2pt] %change the first line to \toprule
			\multirow{3}{*}{\parbox[c]{2cm}{\centering \textbf{Methods}}} & \multicolumn{6}{c|}{\textbf{DN-Wild$\rightarrow$DN-348}} & \multicolumn{6}{c}{\textbf{DN-348$\rightarrow$DN-Wild}}\\
			\cline{2-13} 
			 &\multicolumn{3}{c|}{Day-to-Night } & \multicolumn{3}{c|}{Night-to-Day}& \multicolumn{3}{c|}{Day-to-Night} & \multicolumn{3}{c}{Night-to-Day}  \\
			\cline{2-13}
			 & Rank1 & Rank5 & mAP & Rank1 & Rank5 & mAP & Rank1 & Rank5 & mAP & Rank1 & Rank5 & mAP \\
			% \toprule[2pt] %change the second line to midrule
            \hline
		     PGM~\cite{ref_28}(CVPR'23)&0.515&0.680&0.257&0.634&0.786&0.268&0.388&0.843&0.184&0.365&0.788&0.213\\
                 RPNR~\cite{ref_32}(ACMMM'24) &\underline{0.574}&\underline{0.739}&\underline{0.328}&\underline{0.716}&\underline{0.868}&\underline{0.329}&0.441&0.903&0.204&0.388&0.796&0.208\\
                 PCLHD~\cite{ref_29}(NeurIPS'24) &0.521&0.687&0.261&0.629&0.775&0.257&\underline{0.447}&\underline{0.910}&0.215&\underline{0.418}&\underline{0.865}&\underline{0.236} \\
                 TokenMatcher~\cite{ref_33}(AAAI'25) & 0.480&0.668&0.231&0.597&0.769&0.221&0.413&0.866&0.189&0.376&0.798&0.214\\
                 NULC~\cite{ref_34}(AAAI'25) &0.544&0.707&0.291&0.659&0.801&0.290&0.440&0.909&\underline{0.218}&0.403&0.835&0.222\\
                 PCA~\cite{ref_35}(TIFS'25) &0.533&0.695&0.276&0.635&0.788&0.273&0.433&0.909&0.215&0.405&0.840&0.231\\
                 \hline
                 Ours &\textbf{0.676}&\textbf{0.814}&\textbf{0.375}&\textbf{0.748}&\textbf{0.874}&\textbf{0.375}&\textbf{0.476}&\textbf{0.942}&\textbf{0.240}&\textbf{0.431}&\textbf{0.881}&\textbf{0.254}\\

			\bottomrule[2pt] %change the third line to bottomrule
	\end{tabular}}
    \caption{Zero-shot generalization performance of unsupervised methods. ``DN-Wild$\rightarrow$DN-348'' indicates training on DN-Wild in an unsupervised manner and directly evaluating on DN-348 without fine-tuning.``DN-348$\rightarrow$DN-Wild'' denotes the reverse.}
    \label{table_2}
\end{table*}

\noindent \textbf{Stage-2.} The optimization objective jointly incorporates intra-domain and cross-domain supervision, integrating multiple complementary losses: the pseudo-label-based identity classification loss (${\cal L}_{id}$), the prototype contrastive loss (${\cal L}_{pcl}$), the image-to-text cross-entropy loss (${\cal L}_{i2tce}$), and the cross-domain prototype matching loss (${\cal L}_{cpml}$). The overall stage-2 objective (${\cal L}_{\mathrm{stage2}}$) is formulated as:
\begin{equation}
{\cal L}_{\mathrm{stage2}} = {\cal L}_{id} + {\cal L}_{pcl} + {\cal L}_{i2tce} + {\cal L}_{cpml}.
\label{eq:stage2_loss}
\end{equation}
This design enables the framework to leverage both instance-level semantic alignment and prototype-level cross-domain associations, ensuring robust identity representation learning under the USL-DN-ReID setting.
%Only the visual encoder is updated.
\section{Experiments}
\label{sec:Experiments}

\subsection{Experimental Settings}
\quad\textbf{Datasets.} We evaluate our approach on two publicly available day–night vehicle ReID benchmarks: DN-Wild~\cite{ref_2} and DN-348~\cite{ref_2}. DN-Wild, derived from VERI-Wild 2.0~\cite{ref_37}, exhibits a natural imbalance between day and night images. In contrast, DN-348 provides a balanced set, containing approximately 50 daytime and 50 nighttime images per vehicle ID, collected from a large-scale surveillance system.

\textbf{Evaluation Metrics.} Following established protocols, we evaluate model performance using Cumulative Matching Characteristics (CMC) and mean Average Precision (mAP). In the Day-to-Night (D2N) setting, daytime images serve as queries to retrieve corresponding identities from the nighttime gallery. Conversely, in the Night-to-Day (N2D) setting, nighttime images are used as queries to identify matching identities in the daytime gallery.

\textbf{Implementation Details.} We employ a pre-trained CLIP-B/16 as the backbone for both image and text encoders, with all input images resized to $256 \times 256$. In Stage-1, the model is trained for 150 epochs with a batch size of 64, optimized using Adam~\cite{ref_38} with an initial learning rate of $3.5 \times 10^{-4}$ and a cosine decay schedule, without data augmentation. Stage-2 training is conducted for 40 epochs (300 iterations per epoch) with a weight decay of $5 \times 10^{-4}$, incorporating standard data augmentation including random horizontal flipping, padding, and random erasing~\cite{ref_39}. For clustering, we apply DBSCAN with a maximum distance threshold of 0.7 on DN-348 and 0.8 on DN-Wild, and a minimum of 4 samples per cluster for both datasets. The prototype momentum $\alpha$ and the number of learnable prompts $M$ are set to 0.2 and 4, respectively. All experiments are performed on a single NVIDIA RTX 5090 GPU.

\begin{table*}[!h]
	\belowrulesep=0pt
	\aboverulesep=0pt
	\centering
	\resizebox{\linewidth}{!}{
		\begin{tabular}{c|ccc|ccc|ccc|ccc|ccc}
			\toprule[2pt] %change the first line to \toprule
			\multirow{3}{*}{\centering \textbf{Index}} & \multicolumn{3}{c|}{\multirow{2}{*}{\centering \textbf{Components}}} & \multicolumn{6}{c|}{\textbf{DN-348}} & \multicolumn{6}{c}{\textbf{DN-Wild}}\\
			\cline{5-16} 
			 &&&&\multicolumn{3}{c|}{Day-to-Night } & \multicolumn{3}{c|}{Night-to-Day}& \multicolumn{3}{c|}{Day-to-Night} & \multicolumn{3}{c}{Night-to-Day}  \\
			\cline{2-16}
			& Baseline&IPL&CPML& Rank1 & Rank5 & mAP & Rank1 & Rank5 & mAP & Rank1 & Rank5 & mAP & Rank1 & Rank5 & mAP \\
			% \toprule[2pt] %change the second line to midrule
            \hline
            1& \Checkmark && &0.645 & 0.812 &0.372 & 0.723 &0.863 &0.383 &0.473&0.946&0.261&0.443&0.895&0.277\\
            2& \Checkmark &\Checkmark&&\underline{0.685}&\textbf{0.825}&\underline{0.419}&\underline{0.776}&\textbf{0.906}&\underline{0.423}&\underline{0.492}&0.957&\underline{0.278}&\underline{0.462}&\underline{0.921}&\underline{0.289}\\
            3& \Checkmark &&\Checkmark&0.677&0.808&0.411&0.744&0.878&0.420&0.491&\textbf{0.960}&0.277&0.455&0.906&0.287\\
            4& \Checkmark &\Checkmark&\Checkmark&\textbf{0.707} & \underline{0.823} & \textbf{0.423} & \textbf{0.796} & \underline{0.898} & \textbf{0.437} & \textbf{0.499} & \underline{0.958} & \textbf{0.282} & \textbf{0.476} & \textbf{0.930} & \textbf{0.295}\\
			\bottomrule[2pt] %change the third line to bottomrule
	\end{tabular}}
    \caption{Ablation study of framework components on DN-348 and DN-Wild. IPL and CPML denote the instance-aware prompt learning and cross-domain prototype matching learning modules, respectively.}
    \label{table_3}
\end{table*}

\begin{table}[!h]
	\belowrulesep=0pt
	\aboverulesep=0pt
	\centering
	\resizebox{\linewidth}{!}{
		\begin{tabular}{l|cc|cc}
			\toprule[2pt] %change the first line to \toprule
			\multirow{2}{*}{\centering \textbf{Methods}} & \multicolumn{2}{c|}{Day-to-Night } & \multicolumn{2}{c}{Night-to-Day}\\
			\cline{2-5}
			& Rank1 & mAP &Rank1 & mAP \\
			% \toprule[2pt] %change the second line to midrule
            \hline
            \textbf{Ours (Two-stage)}& \textbf{0.707} & \textbf{0.423} & \textbf{0.796} & \textbf{0.437} \\
            \hline
            w/o Stage-1& 0.666&0.402&0.752&0.415\\
            w/o Dynamic-bias Net& 0.672&0.415&\underline{0.776}&\underline{0.428}\\
            Stage-2 w/ Instance Align&\underline{0.682}&\underline{0.417}&0.772&\underline{0.428} \\
			\bottomrule[2pt] %change the third line to bottomrule
	\end{tabular}}
    \caption{Ablation study on textual prompt learning on DN-348.}
    \label{table_4}
\end{table}

\begin{table}[!h]
	\belowrulesep=0pt
	\aboverulesep=0pt
	\centering
	\resizebox{\linewidth}{!}{
		\begin{tabular}{l|cc|cc|cc|cc}
			\toprule[2pt] %change the first line to \toprule
			\multirow{3}{*}{\centering \textbf{M}} & \multicolumn{4}{c|}{\textbf{DN-348}}& \multicolumn{4}{c}{\textbf{DN-Wild}}\\
			\cline{2-9} 
			 &\multicolumn{2}{c|}{Day-to-Night } & \multicolumn{2}{c|}{Night-to-Day}&\multicolumn{2}{c|}{Day-to-Night } & \multicolumn{2}{c}{Night-to-Day}\\
			\cline{2-9}
			& Rank1 & mAP &Rank1 & mAP & Rank1 & mAP &Rank1 & mAP\\
			% \toprule[2pt] %change the second line to midrule
            \hline
            1&0.692&0.417&0.785&\underline{0.433}&\underline{0.497} & 0.281 &\underline{0.472} &\underline{0.292}\\
            2&\underline{0.698}&\underline{0.420}&\underline{0.791}&0.431&0.495&\textbf{0.284}&\underline{0.472}&0.290\\
            4&\textbf{0.707} & \textbf{0.423} & \textbf{0.796} & \textbf{0.437} &\textbf{0.499}&\underline{0.282}&\textbf{0.476}&\textbf{0.295}\\
            6&0.687&0.419&0.783&0.430 & 0.494 & 0.280 & 0.468&0.288\\
            8&0.678&0.411&0.745&0.421 & 0.488&0.278&0.465&0.288\\
			\bottomrule[2pt] %change the third line to bottomrule
	\end{tabular}}
    \caption{Performance analysis on the number of prompts M.}
    \label{table_5}
    \vspace{-6pt}
\end{table}
\subsection{Comparison with State-of-the-Art Methods}
We evaluate the effectiveness of the proposed framework through comprehensive comparisons with state-of-the-art approaches, summarized in Tables~\ref{table_1} and \ref{table_2}. As no prior methods are specifically designed for USL-DN-ReID, we adopt representative USL-VI-ReID methods—capable of learning without manual annotations—as baselines.

\textbf{USL-DN-ReID Performance.}
As reported in Table~\ref{table_1}, our method achieves state-of-the-art performance across all benchmark settings. On DN-348, it attains Rank-1 accuracies of 70.7\% and 79.6\% under the two cross-domain retrieval scenarios, substantially surpassing existing USL-VI-ReID baselines. These results indicate that prior methods struggle with the pronounced day–night domain gap. Moreover, our approach remains highly competitive relative to fully supervised learning (FSL) methods, highlighting its effectiveness and robustness in unsupervised settings.

\textbf{Zero-Shot Generalization.}
Table~\ref{table_2} presents the zero-shot cross-domain performance of USL methods. Our framework consistently achieves strong generalization across datasets, maintaining comparable accuracy to direct testing results in Table~\ref{table_1} without noticeable performance degradation. This demonstrates the model’s capacity to learn domain-invariant representations, enabling reliable association of daytime and nighttime identities under unseen conditions. This validates both the superior performance and robust zero-shot generalization of our method.

\subsection{Ablation Study}
\textbf{Baseline Model.}
Our baseline, denoted as IIA, is obtained by removing both the instance-aware prompt learning (IPL) module and the cross-domain prototype matching learning (CPML) module. Consequently, the model is optimized solely with the intra-domain identity association losses $\mathcal{L}_{id} + \mathcal{L}_{pcl}$.

\textbf{Effect of Individual Modules.}
We evaluate the contribution of each component, with results summarized in Table~\ref{table_3}. The baseline model leveraging only IIA achieves reasonable performance across all metrics, demonstrating the efficacy of prototype contrastive learning for USL-DN-ReID. Integrating the IPL module substantially improves results; for instance, on DN-348 under Day-to-Night retrieval, Rank-1 and mAP increase by 4.0\% and 4.7\%, respectively. This confirms that instance-level textual prompts provide effective semantic supervision, enhancing feature discriminability. Similarly, incorporating the CPML module independently yields notable performance gains, indicating that explicit mining of positive and negative cross-domain prototype pairs effectively establishes identity correspondences. Combining both IPL and CPML further boosts performance, illustrating their complementarity.

\textbf{Impact of Textual Prompt Learning Strategy.}
We investigate the effect of the stage-1 training and dynamic prompt adaptation (Table~\ref{table_4}). Omitting stage-1 results in substantial performance degradation, highlighting the necessity of initial vision-language alignment for effective unsupervised learning in stage-2. Additionally, ablating the dynamic-bias network or freezing instance-level alignment in stage-2 leads to reduced accuracy. These findings validate that static prompt vectors are insufficient to accommodate dynamically evolving pseudo-label distributions, and underscore the importance of our two-stage progressive strategy with dynamic prompt adaptation.

\textbf{Number of Learnable Prompts.}
We analyze the sensitivity of the model to the number of learnable prompt vectors $M$ in the IPL module (Table~\ref{table_5}). Results indicate moderate sensitivity: an appropriate number of prompts enhances the diversity and discriminability of textual descriptions, aiding visual–textual alignment, whereas excessive prompts introduce redundancy and noise, impairing generalization. Based on these observations, we adopt $M=4$.

\textbf{Cross-Domain Top-$k$ Neighbors.}
We evaluate the influence of the top-$k$ parameter in CPML prototype matching (Figure~\ref{img_3}). Performance remains stable across a moderate range of $k$, but degrades when $k$ is excessively large, likely due to noisy or ambiguous prototype associations. Optimal performance is achieved with $k=20$ and $k=15$ for DN-Wild and DN-348, respectively.

\begin{figure}[!h]
    \centering
    \includegraphics[width=.49\textwidth]{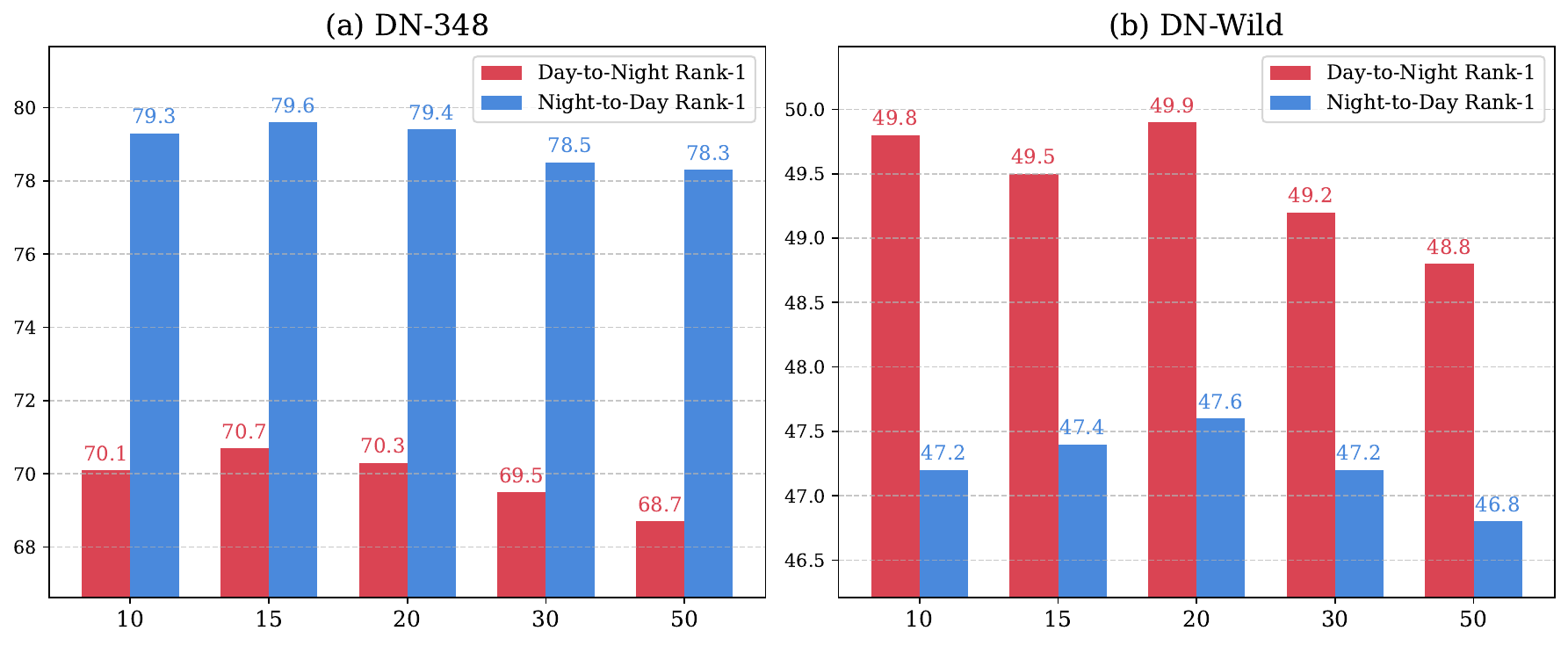}
    \caption{Impact of the number of cross-domain top-$k$ neighbors on model performance.}
    \label{img_3}
    \vspace{-5pt}
\end{figure}

\subsection{Visualization}

% \textbf{Cosine Similarity Distributions of Positive-Negative Pairs.} To quantitatively evaluate the feature discriminability, we present the cosine similarity distributions of randomly sampled day-night image pairs from the DN-348 dataset. A comparison between the baseline and our method in Figures \ref{img_4}(a-b) reveals that our approach substantially sharpens the cross-domain similarity distribution: (i) the density of positive pairs becomes more concentrated at higher similarity values, and (ii) the margin $\delta$ between the means of positive and negative distributions expands from 0.200 to 0.261. This demonstrates a stronger alignment of cross-domain positives and a clearer separation from negatives, which is pivotal for reliable cross-domain identity association.

\begin{figure}[!h]
    \centering
    \includegraphics[width=.49\textwidth]{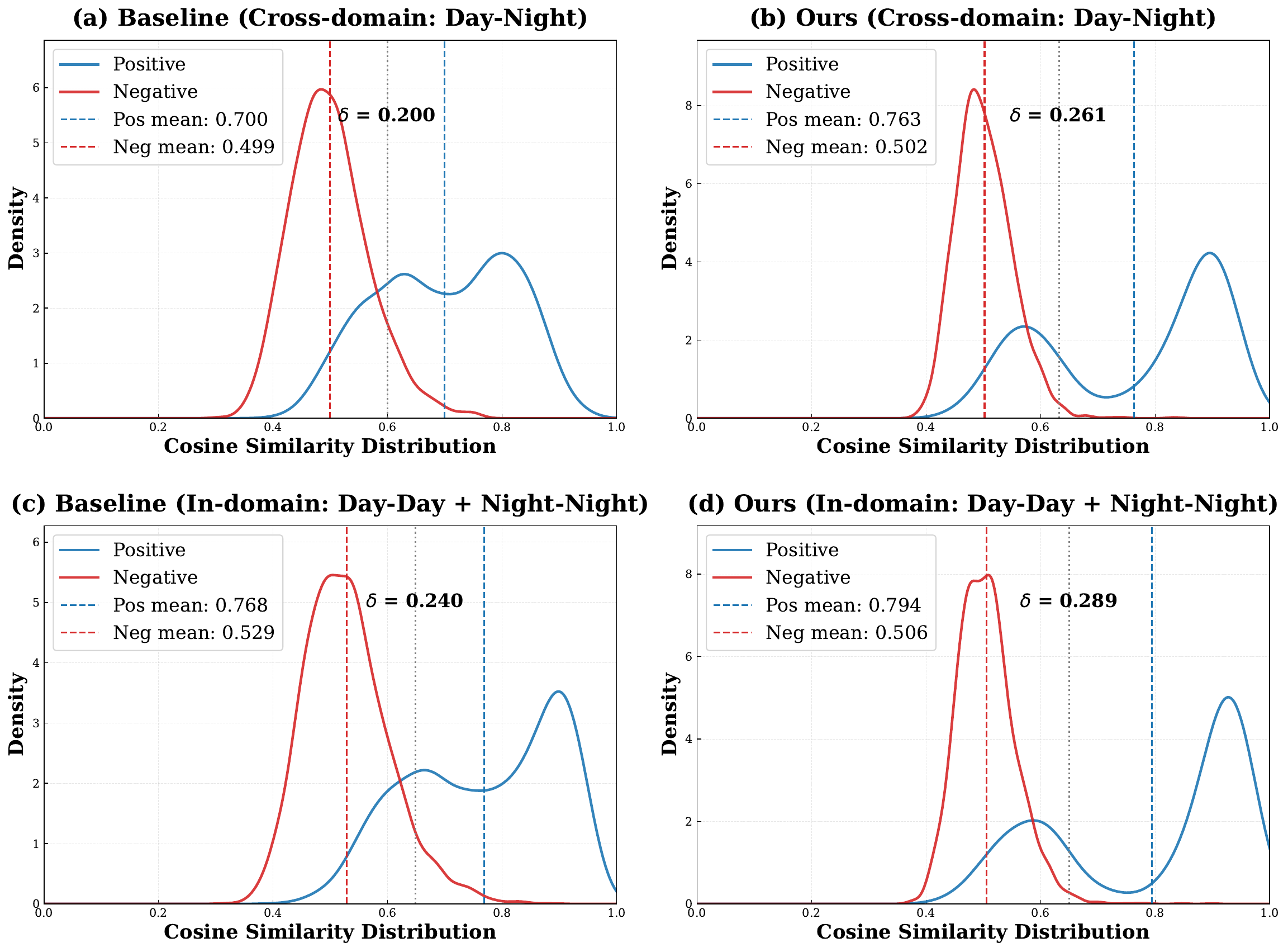}
    \caption{Cosine similarity distributions of positive and negative day–night pairs on the DN-348 dataset. $\delta$ denotes the margin between the mean similarities of positive and negative pairs.}
    \label{img_4}
    \vspace{-4pt}
\end{figure}

\textbf{Cosine Similarity Distributions of Positive and Negative Pairs.}
To quantitatively assess feature discriminability, we visualize the cosine similarity distributions of randomly sampled day–night pairs from DN-348. As shown in Figures~\ref{img_4} (a–b), our method significantly sharpens the cross-domain similarity distribution: (i) positive pairs concentrate at higher similarity values, and (ii) the mean margin $\delta$ between positive and negative pairs increases from 0.200 to 0.261. This indicates enhanced alignment of cross-domain positives and clearer separation from negatives, which is crucial for reliable identity association.
Figures~\ref{img_4} (c–d) further show analogous improvements for in-domain pairs. Collectively, these results demonstrate that our framework effectively enforces both cross-domain identity consistency and intra-domain feature discriminability.

% \textbf{Cosine Similarity Distributions.} we visualize the cosine similarity distributions of randomly sampled day-night image pairs from the DN-348 dataset. Specifically, Figures \ref{img_5} (a) and (b) compare the similarity distributions of cross-domain positive and negative pairs between the baseline and our method. Two key observations can be made: (1) our method exhibits a notably higher density concentration around the positive similarity region, and (2) the margin $\delta$ between the means of cross-domain positive and negative distributions increases from 0.200 to 0.261. These findings indicate that our approach achieves stronger alignment of cross-domain positives and greater separation from negatives, leading to more reliable identity association across domains.

% Furthermore, Figures \ref{img_5} (c) and (d) depict the distributions of in-domain positive and negative pairs, where our method also demonstrates a clear improvement in separability between positive and negative pairs. Collectively, these results emphasize that the proposed framework not only enhances cross-domain identity consistency but also strengthens feature discriminability within each domain.

\begin{figure}[!h]
    \centering
    \includegraphics[width=.49\textwidth]{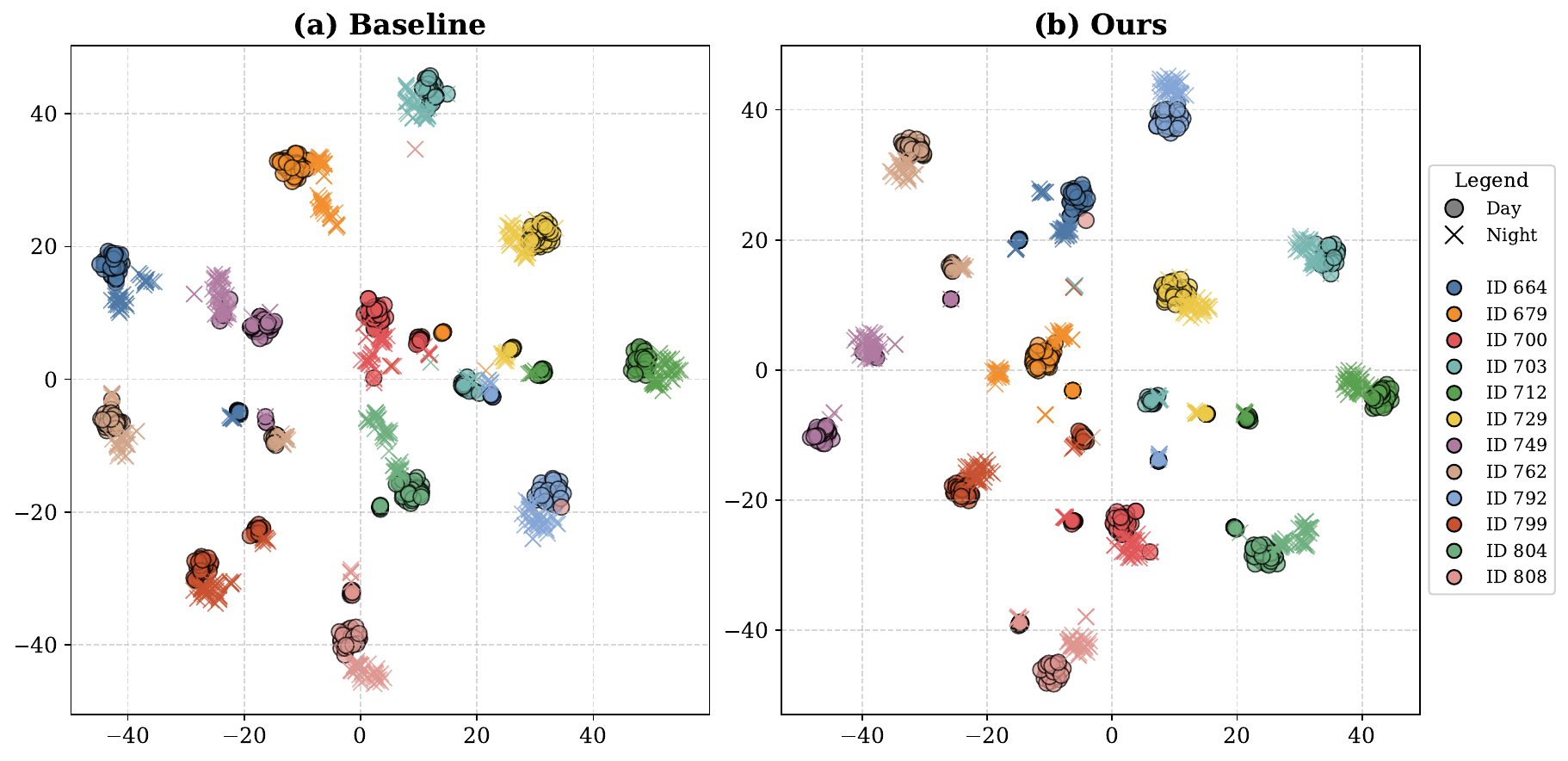}
    \caption{t-SNE visualization of cross-domain feature distributions on DN-348. The same color indicates the same identity.}
    \label{img_5}
    \vspace{-6pt}
\end{figure}

\noindent\textbf{Cross-Domain t-SNE Embeddings.}
We visualize cross-domain feature embeddings using t-SNE \cite{ref_42} in Figure~\ref{img_5}. While the baseline exhibits moderate class separation, our method produces more compact intra-class clusters and clearer inter-class boundaries. These results confirm that the proposed framework enhances cross-domain feature discriminability, facilitating reliable identity association.

% that the proposed framework not only enhances cross-domain identity consistency but also strengthens feature discriminability within each domain.that the proposed framework not only enhances cross-domain identity consistency but also strengthens feature discriminability within each domainthat the proposed framework not only enhances cross-domain identity consistency

\section{Conclusion}
\label{sec:Conclusion}
We propose a novel framework for unsupervised day–night vehicle re-identification (USL-DN-ReID) that synergistically combines instance-aware prompt learning with cross-domain prototype consistency modeling. Leveraging the semantic alignment capabilities of vision–language pre-trained models, our approach addresses the challenges of large illumination discrepancies, absence of identity annotations, and complex cross-domain identity correspondences.
Specifically, in Stage-1, we introduce a dynamic-bias instance-aware prompt learning module that generates adaptive textual descriptions for unlabeled images, establishing fine-grained image–text correspondence without requiring identity supervision. In Stage-2, we employ intra-domain identity association to enhance feature discriminability within domains via prototype memory banks, and cross-domain prototype matching to reliably mine positive and negative prototype pairs across day and night domains, thereby achieving robust cross-domain identity alignment.
Extensive experiments on DN-348 and DN-Wild benchmarks demonstrate that our framework achieves state-of-the-art performance under unsupervised settings, surpassing existing USL methods and remaining competitive with fully supervised approaches. 

\noindent\textbf{Broader Impact:} Our framework exhibits strong zero-shot generalization across unseen domains and holds promise for building scalable and annotation-efficient ReID systems. By advancing unsupervised cross-domain learning, it has the potential to reduce reliance on large-scale manual labeling and facilitate deployment in diverse real-world environment.

\section*{Acknowledgments}
This work is supported by the National Natural Science Foundation of China under Grants (62571380), and the Hubei Technical Innovation Program (2025BEA002).
{
    \small
    \bibliographystyle{ieeenat_fullname}
    \bibliography{main}

@String(CVPR= {IEEE Conf. Comput. Vis. Pattern Recog.})

@String(ICCV= {Int. Conf. Comput. Vis.})

@String(ECCV= {Eur. Conf. Comput. Vis.})

@String(NIPS= {Adv. Neural Inform. Process. Syst.})

@String(AAAI = {AAAI})

@String(CVPR  = {CVPR})

@String(ICCV  = {ICCV})

@String(ECCV  = {ECCV})

@String(NIPS  = {NeurIPS})

@ARTICLE{ref_1,
  author={Lu, Andong and Li, Chenglong and Zha, Tianrui and Wang, Xiao-Feng and Tang, Jin and Luo, Bin},
  journal={IEEE Transactions on Information Forensics and Security}, 
  title={Nighttime Person Re-Identification via Collaborative Enhancement Network With Multi-Domain Learning}, 
  year={2025},
  volume={20},
  number={},
  pages={1305-1319},
  doi={10.1109/TIFS.2025.3527335}}

@INPROCEEDINGS{ref_2,
  author={Li, Hongchao and Chen, Jingong and Zheng, Aihua and Wu, Yong and Luo, Yonglong},
  booktitle={2024 IEEE/CVF Conference on Computer Vision and Pattern Recognition (CVPR)}, 
  title={Day-Night Cross-domain Vehicle Re-identification}, 
  year={2024},
  volume={},
  number={},
  pages={12626-12635},
  doi={10.1109/CVPR52733.2024.01200}}

@article{ref_4,
title={Learning Progressive Modality-Shared Transformers for Effective Visible-Infrared Person Re-identification}, 
volume={37}, 
url={https://ojs.aaai.org/index.php/AAAI/article/view/25273}, 
DOI={10.1609/aaai.v37i2.25273}, 
number={2},
journal={Proceedings of the AAAI Conference on Artificial Intelligence},
author={Lu, Hu and Zou, Xuezhang and Zhang, Pingping}, year={2023}, month={Jun.}, pages={1835-1843} }

@inproceedings{ref_5,
author = {Sun, Hanzhe and Liu, Jun and Zhang, Zhizhong and Wang, Chengjie and Qu, Yanyun and Xie, Yuan and Ma, Lizhuang},
title = {Not All Pixels Are Matched: Dense Contrastive Learning for Cross-Modality Person Re-Identification},
year = {2022},
isbn = {9781450392037},
publisher = {Association for Computing Machinery},
address = {New York, NY, USA},
url = {https://doi.org/10.1145/3503161.3547970},
doi = {10.1145/3503161.3547970},
booktitle = {Proceedings of the 30th ACM International Conference on Multimedia},
pages = {5333–5341},
numpages = {9},
location = {Lisboa, Portugal},
series = {MM '22}
}

@InProceedings{ref_6,
  title = 	 {{BLIP}: Bootstrapping Language-Image Pre-training for Unified Vision-Language Understanding and Generation},
  author =       {Li, Junnan and Li, Dongxu and Xiong, Caiming and Hoi, Steven},
  booktitle = 	 {Proceedings of the 39th International Conference on Machine Learning},
  pages = 	 {12888--12900},
  year = 	 {2022},
  editor = 	 {Chaudhuri, Kamalika and Jegelka, Stefanie and Song, Le and Szepesvari, Csaba and Niu, Gang and Sabato, Sivan},
  volume = 	 {162},
  series = 	 {Proceedings of Machine Learning Research},
  month = 	 {17--23 Jul},
  publisher =    {PMLR},
  url = 	 {https://proceedings.mlr.press/v162/li22n.html},
}

@inproceedings{ref_8,
    title =      {Learning Transferable Visual Models From Natural Language Supervision},
    author =       {Radford, Alec and Kim, Jong Wook and Hallacy, Chris and Ramesh, Aditya and Goh, Gabriel and Agarwal, Sandhini and Sastry, Girish and Askell, Amanda and Mishkin, Pamela and Clark, Jack and Krueger, Gretchen and Sutskever, Ilya},
    booktitle =      {Proceedings of the 38th International Conference on Machine Learning},
    pages =      {8748--8763},
    year =      {2021},
    volume =      {139},
    month =      {18--24 Jul},
    url =      {https://proceedings.mlr.press/v139/radford21a.html},
}

@article{ref_9,
title={CLIP-ReID: Exploiting Vision-Language Model for Image Re-identification without Concrete Text Labels},
volume={37},
number={1},
journal={Proceedings of the AAAI Conference on Artificial Intelligence},
author={Li, Siyuan and Sun, Li and Li, Qingli},
year={2023},
month={Jun.},
pages={1405-1413} }

@article{ref_10,
    title={Learning to Prompt for Vision-Language Models},
    author={Zhou, Kaiyang and Yang, Jingkang and Loy, Chen Change and Liu, Ziwei},
    journal={International Journal of Computer Vision},
    year={2022}
}

@article{ref_11, 
title={Multi-Prompts Learning with Cross-Modal Alignment for Attribute-Based Person Re-identification},
volume={38},
url={https://ojs.aaai.org/index.php/AAAI/article/view/28524}, 
DOI={10.1609/aaai.v38i7.28524},
number={7},
journal={Proceedings of the AAAI Conference on Artificial Intelligence},
author={Zhai, Yajing and Zeng, Yawen and Huang, Zhiyong and Qin, Zheng and Jin, Xin and Cao, Da}, year={2024}, month={Mar.}, pages={6979-6987} }

@ARTICLE{ref_12,
  author={He, Shuting and Chen, Weihua and Wang, Kai and Luo, Hao and Wang, Fan and Jiang, Wei and Ding, Henghui},
  journal={IEEE Transactions on Information Forensics and Security}, 
  title={Region Generation and Assessment Network for Occluded Person Re-Identification}, 
  year={2024},
  volume={19},
  number={},
  pages={120-132},
  doi={10.1109/TIFS.2023.3318956}}

@ARTICLE{ref_13,
  author={Ye, Mang and Wu, Zesen and Du, Bo},
  journal={IEEE Transactions on Pattern Analysis and Machine Intelligence}, 
  title={Dual-Level Matching With Outlier Filtering for Unsupervised Visible-Infrared Person Re-Identification}, 
  year={2025},
  volume={47},
  number={5},
  pages={3815-3829},
  doi={10.1109/TPAMI.2025.3541053}}

@ARTICLE{ref_14,
  author={Li, Zhiyong and Liu, Haojie and Peng, Xiantao and Jiang, Wei},
  journal={IEEE Transactions on Knowledge and Data Engineering}, 
  title={Inter-Intra Modality Knowledge Learning and Clustering Noise Alleviation for Unsupervised Visible-Infrared Person Re-Identification}, 
  year={2024},
  volume={36},
  number={8},
  pages={3934-3947},
  doi={10.1109/TKDE.2024.3367304}}

@INPROCEEDINGS{ref_15,
  author={Yang, Bin and Chen, Jun and Ye, Mang},
  booktitle={2024 IEEE/CVF Conference on Computer Vision and Pattern Recognition (CVPR)}, 
  title={Shallow-Deep Collaborative Learning for Unsupervised Visible-Infrared Person Re-Identification}, 
  year={2024},
  volume={},
  number={},
  pages={16870-16879},
  doi={10.1109/CVPR52733.2024.01596}}

@InProceedings{ref_16,
author="Shi, Jiangming
and Yin, Xiangbo
and Chen, Yeyun
and Zhang, Yachao
and Zhang, Zhizhong
and Xie, Yuan
and Qu, Yanyun",
title="Multi-memory Matching for Unsupervised Visible-Infrared Person Re-identification",
booktitle="Computer Vision -- ECCV 2024",
year="2025",
pages="456--474",
}

@ARTICLE{ref_17,
  author={Zhang, Yi-Feng and Zhang, Can-Long and Tian, Jun-Wei and Ma, Hai-Fei and Li, Zhi-Xin and Wang, Zhi-Wen},
  journal={IEEE Transactions on Circuits and Systems for Video Technology}, 
  title={CMAG: Cross-Modal Attention and Graph-Enhanced Memory for Unsupervised Visible-Infrared Person Re-Identification}, 
  year={2025},
  volume={},
  number={},
  pages={1-1},
  doi={10.1109/TCSVT.2025.3595846}}

@ARTICLE{ref_18,
  author={Pang, Zhiqi and Wang, Chunyu and Zhao, Lingling and Liu, Yang and Sharma, Gaurav},
  journal={IEEE Transactions on Circuits and Systems for Video Technology}, 
  title={Cross-Modality Hierarchical Clustering and Refinement for Unsupervised Visible-Infrared Person Re-Identification}, 
  year={2024},
  volume={34},
  number={4},
  pages={2706-2718},
  doi={10.1109/TCSVT.2023.3310015}}

@inproceedings{ref_19,
 author = {Miyai, Atsuyuki and Yu, Qing and Irie, Go and Aizawa, Kiyoharu},
 booktitle = {Advances in Neural Information Processing Systems},
 editor = {A. Oh and T. Naumann and A. Globerson and K. Saenko and M. Hardt and S. Levine},
 pages = {76298--76310},
 publisher = {Curran Associates, Inc.},
 title = {LoCoOp: Few-Shot Out-of-Distribution Detection via Prompt Learning},
 url = {https://proceedings.neurips.cc/paper_files/paper/2023/file/f0606b882692637835e8ac981089eccd-Paper-Conference.pdf},
 volume = {36},
 year = {2023}
}

@inproceedings{ref_20,
author = {Wu, Jiannan and Zhong, Muyan and Xing, Sen and Lai, Zeqiang and Liu, Zhaoyang and Chen, Zhe and Wang, Wenhai and Zhu, Xizhou and Lu, Lewei and Lu, Tong and Luo, Ping and Qiao, Yu and Dai, Jifeng},
title = {VisionLLM v2: an end-to-end generalist multimodal large language model for hundreds of vision-language tasks},
year = {2024},
isbn = {9798331314385},
booktitle = {Proceedings of the 38th International Conference on Neural Information Processing Systems},
articleno = {2235},
numpages = {51},
location = {Vancouver, BC, Canada},
series = {NIPS '24}
}

@article{ref_21, title={CLIP-driven View-aware Prompt Learning for Unsupervised Vehicle Re-identification}, volume={39}, url={https://ojs.aaai.org/index.php/AAAI/article/view/32962}, DOI={10.1609/aaai.v39i8.32962}, number={8}, journal={Proceedings of the AAAI Conference on Artificial Intelligence}, author={Xu, Jiyang and Wang, Qi and Xiong, Xin and Gai, Di and Zhou, Ruihua and Wang, Dong}, year={2025}, month={Apr.}, pages={8896-8904} }

@ARTICLE{ref_22,
  author={Yu, Xiaoyan and Dong, Neng and Zhu, Liehuang and Peng, Hao and Tao, Dapeng},
  journal={IEEE Transactions on Multimedia}, 
  title={CLIP-Driven Semantic Discovery Network for Visible-Infrared Person Re-Identification}, 
  year={2025},
  volume={27},
  number={},
  pages={4137-4150},
  doi={10.1109/TMM.2025.3535353}}

@ARTICLE{ref_23,
  author={Zhang, Guoqing and Zhou, Jieqiong and Jiang, Lu and Zheng, Yuhui and Lin, Weisi},
  journal={IEEE Transactions on Image Processing}, 
  title={CLIP-Based Multi-Modal Feature Learning for Cloth-Changing Person Re-Identification}, 
  year={2025},
  volume={34},
  number={},
  pages={5570-5583},
  doi={10.1109/TIP.2025.3602641}}

@ARTICLE{ref_24,
  author={Liang, Wenqi and Wang, Guangcong and Lai, Jianhuang and Xie, Xiaohua},
  journal={IEEE Transactions on Image Processing}, 
  title={Homogeneous-to-Heterogeneous: Unsupervised Learning for RGB-Infrared Person Re-Identification}, 
  year={2021},
  volume={30},
  number={},
  pages={6392-6407},
  doi={10.1109/TIP.2021.3092578}}

@article{ref_25,
title={NightReID: A Large-Scale Nighttime Person Re-Identification Benchmark},
volume={39},
url={https://ojs.aaai.org/index.php/AAAI/article/view/33142},
DOI={10.1609/aaai.v39i10.33142},
number={10}, journal={Proceedings of the AAAI Conference on Artificial Intelligence}, author={Zhao, Yuxuan and Ruan, Weijian and Li, He and Ye, Mang}, year={2025}, month={Apr.}, pages={10519-10527} }

@INPROCEEDINGS{ref_26,
  author={He, Kaiming and Fan, Haoqi and Wu, Yuxin and Xie, Saining and Girshick, Ross},
  booktitle={2020 IEEE/CVF Conference on Computer Vision and Pattern Recognition (CVPR)}, 
  title={Momentum Contrast for Unsupervised Visual Representation Learning}, 
  year={2020},
  volume={},
  number={},
  pages={9726-9735},
  doi={10.1109/CVPR42600.2020.00975}}

@inproceedings{ref_27,
author = {Snell, Jake and Swersky, Kevin and Zemel, Richard},
title = {Prototypical networks for few-shot learning},
year = {2017},
isbn = {9781510860964},
booktitle = {Proceedings of the 31st International Conference on Neural Information Processing Systems},
pages = {4080–4090},
numpages = {11},
series = {NIPS'17}
}

@INPROCEEDINGS{ref_28,
  author={Wu, Zesen and Ye, Mang},
  booktitle={2023 IEEE/CVF Conference on Computer Vision and Pattern Recognition (CVPR)}, 
  title={Unsupervised Visible-Infrared Person Re-Identification via Progressive Graph Matching and Alternate Learning}, 
  year={2023},
  volume={},
  number={},
  pages={9548-9558},
  doi={10.1109/CVPR52729.2023.00921}}

@inproceedings{ref_29,
author = {Shi, Jiangming and Yin, Xiangbo and Zhang, Yachao and Zhang, Zhizhong and Xie, Yuan and Qu, Yanyun},
title = {Learning commonality, divergence and variety for unsupervised visible-infrared person re-identification},
year = {2024},
isbn = {9798331314385},
booktitle = {Proceedings of the 38th International Conference on Neural Information Processing Systems},
articleno = {3163},
numpages = {20},
series = {NIPS '24}
}

@INPROCEEDINGS{ref_30,
  author={Ye, Mang and Ruan, Weijian and Du, Bo and Shou, Mike Zheng},
  booktitle={2021 IEEE/CVF International Conference on Computer Vision (ICCV)}, 
  title={Channel Augmented Joint Learning for Visible-Infrared Recognition}, 
  year={2021},
  volume={},
  number={},
  pages={13547-13556},
  doi={10.1109/ICCV48922.2021.01331}}

@ARTICLE{ref_31,
  author={Ye, Mang and Shen, Jianbing and Lin, Gaojie and Xiang, Tao and Shao, Ling and Hoi, Steven C. H.},
  journal={IEEE Transactions on Pattern Analysis and Machine Intelligence}, 
  title={Deep Learning for Person Re-Identification: A Survey and Outlook}, 
  year={2022},
  volume={44},
  number={6},
  pages={2872-2893},
  doi={10.1109/TPAMI.2021.3054775}}

@inproceedings{ref_32,
author = {Yin, Xiangbo and Shi, Jiangming and Zhang, Yachao and Lu, Yang and Zhang, Zhizhong and Xie, Yuan and Qu, Yanyun},
title = {Robust Pseudo-label Learning with Neighbor Relation for Unsupervised Visible-Infrared Person Re-Identification},
year = {2024},
isbn = {9798400706868},
url = {https://doi.org/10.1145/3664647.3680951},
doi = {10.1145/3664647.3680951},
booktitle = {Proceedings of the 32nd ACM International Conference on Multimedia},
pages = {2242–2251},
numpages = {10},
location = {Melbourne VIC, Australia},
series = {MM '24}
}

@article{ref_33, 
title={TokenMatcher: Diverse Tokens Matching for Unsupervised Visible-Infrared Person Re-Identification},
volume={39}, url={https://ojs.aaai.org/index.php/AAAI/article/view/32855},
DOI={10.1609/aaai.v39i8.32855},
number={8},
journal={Proceedings of the AAAI Conference on Artificial Intelligence},
author={Wang, Xiao and Liu, Lekai and Yang, Bin and Ye, Mang and Wang, Zheng and Xu, Xin},
year={2025}, month={Apr.}, pages={7934-7942} }

@article{ref_34,
title={Relieving Universal Label Noise for Unsupervised Visible-Infrared Person Re-Identification by Inferring from Neighbors},
volume={39},
url={https://ojs.aaai.org/index.php/AAAI/article/view/32791},
DOI={10.1609/aaai.v39i7.32791},
number={7},
journal={Proceedings of the AAAI Conference on Artificial Intelligence},
author={Teng, Xiao and Lan, Long and Chen, Dingyao and Xu, Kele and Yin, Nan}, year={2025}, month={Apr.}, pages={7356-7364} }

@ARTICLE{ref_35,
  author={Yang, Yiming and Hu, Weipeng and Hu, Haifeng},
  journal={IEEE Transactions on Information Forensics and Security}, 
  title={Progressive Cross-Modal Association Learning for Unsupervised Visible-Infrared Person Re-Identification}, 
  year={2025},
  volume={20},
  number={},
  pages={1290-1304},
  doi={10.1109/TIFS.2025.3527356}}

@INPROCEEDINGS{ref_36,
  author={Zhou, Kaiyang and Yang, Jingkang and Loy, Chen Change and Liu, Ziwei},
  booktitle={2022 IEEE/CVF Conference on Computer Vision and Pattern Recognition (CVPR)}, 
  title={Conditional Prompt Learning for Vision-Language Models}, 
  year={2022},
  volume={},
  number={},
  pages={16795-16804},
  doi={10.1109/CVPR52688.2022.01631}}

@ARTICLE{ref_37,
  author={Bai, Yan and Liu, Jun and Lou, Yihang and Wang, Ce and Duan, Ling-Yu},
  journal={IEEE Transactions on Pattern Analysis and Machine Intelligence}, 
  title={Disentangled Feature Learning Network and a Comprehensive Benchmark for Vehicle Re-Identification}, 
  year={2022},
  volume={44},
  number={10},
  pages={6854-6871},
  doi={10.1109/TPAMI.2021.3099253}}

@misc{ref_38,
      title={Adam: A Method for Stochastic Optimization}, 
      author={Diederik P. Kingma and Jimmy Ba},
      year={2017},
      eprint={1412.6980},
      archivePrefix={arXiv},
      primaryClass={cs.LG},
      url={https://arxiv.org/abs/1412.6980}, 
}

@article{ref_39,
title={Random Erasing Data Augmentation},
volume={34},
url={https://ojs.aaai.org/index.php/AAAI/article/view/7000},
DOI={10.1609/aaai.v34i07.7000},
number={07},
journal={Proceedings of the AAAI Conference on Artificial Intelligence},
author={Zhong, Zhun and Zheng, Liang and Kang, Guoliang and Li, Shaozi and Yang, Yi},
year={2020},
month={Apr.},
pages={13001-13008} }

@InProceedings{ref_40,
author="Wei, Haoran
and Kong, Lingyu
and Chen, Jinyue
and Zhao, Liang
and Ge, Zheng
and Yang, Jinrong
and Sun, Jianjian
and Han, Chunrui
and Zhang, Xiangyu",
editor="Leonardis, Ale{\v{s}}
and Ricci, Elisa
and Roth, Stefan
and Russakovsky, Olga
and Sattler, Torsten
and Varol, G{\"u}l",
title="Vary: Scaling up the Vision Vocabulary for Large Vision-Language Model",
booktitle="Computer Vision -- ECCV 2024",
year="2025",
pages="408--424",
isbn="978-3-031-73235-5"
}

@inproceedings{ref_41,
author = {Cheng, De and He, Lingfeng and Wang, Nannan and Zhang, Shizhou and Wang, Zhen and Gao, Xinbo},
title = {Efficient Bilateral Cross-Modality Cluster Matching for Unsupervised Visible-Infrared Person ReID},
year = {2023},
isbn = {9798400701085},
url = {https://doi.org/10.1145/3581783.3612073},
doi = {10.1145/3581783.3612073},
booktitle = {Proceedings of the 31st ACM International Conference on Multimedia},
pages = {1325–1333},
numpages = {9},
series = {MM '23}
}

@article{ref_42,
  author  = {Laurens van der Maaten and Geoffrey Hinton},
  title   = {Visualizing Data using t-SNE},
  journal = {Journal of Machine Learning Research},
  year    = {2008},
  volume  = {9},
  number  = {86},
  pages   = {2579--2605},
  url     = {http://jmlr.org/papers/v9/vandermaaten08a.html}
}

@ARTICLE{ref_43,
  author={Liu, Yexin and Zhang, Weiming and Vasilakos, Athanasios V. and Wang, Lin},
  journal={IEEE Transactions on Neural Networks and Learning Systems}, 
  title={Unsupervised Visible–Infrared ReID via Pseudo-Label Correction and Modality-Level Alignment}, 
  year={2025},
  volume={36},
  number={11},
  pages={19631-19643},
  doi={10.1109/TNNLS.2025.3591641}}

@INPROCEEDINGS{ref_44,
  author={Khattak, Muhammad Uzair and Rasheed, Hanoona and Maaz, Muhammad and Khan, Salman and Khan, Fahad Shahbaz},
  booktitle={2023 IEEE/CVF Conference on Computer Vision and Pattern Recognition (CVPR)}, 
  title={MaPLe: Multi-modal Prompt Learning}, 
  year={2023},
  volume={},
  number={},
  pages={19113-19122},
  doi={10.1109/CVPR52729.2023.01832}}

@inproceedings{ref_45,
  title={Consistency-guided Prompt Learning for Vision-Language Models},
  author={Roy, Shuvendu and Etemad, Ali},
  booktitle="International Conference on Learning Representations",
  year={2024}
}

@article{ref_46,
author = {Du, Songcheng and Zou, Yang and Wang, Zixu and Li, Xingyuan and Li, Ying and Shang, Changjing and Shen, Qiang},
title = {Unsupervised Hyperspectral Image Super-Resolution via Self-Supervised Modality Decoupling},
year = {2026},
volume = {134},
number = {4},
issn = {0920-5691},
url = {https://doi.org/10.1007/s11263-026-02757-8},
doi = {10.1007/s11263-026-02757-8},
journal = {Int. J. Comput. Vision},
numpages = {23},
}

@article{ref_48,
  title={Reclaiming Lost Text Layers for Source-Free Cross-Domain Few-Shot Learning},
  author={Zhang, Zhenyu and Chen, Guangyao and Zou, Yixiong and Li, Yuhua and Li, Ruixuan},
  journal={arXiv preprint arXiv:2603.05235},
  year={2026}
}

@InProceedings{ref_49,
    author    = {Wang, Qi and Zhang, Zeyu and Wang, Dong and Gai, Di and Xiong, Xin and Xu, Jiyang and Zhou, Ruihua},
    title     = {VehicleMAE: View-asymmetry Mutual Learning for Vehicle Re-identification Pre-training via Masked AutoEncoders},
    booktitle = {Proceedings of the IEEE/CVF International Conference on Computer Vision (ICCV)},
    month     = {October},
    year      = {2025},
    pages     = {4701-4711}
}
}

\clearpage
\setcounter{page}{1}
\maketitlesupplementary

In this supplementary material, we present further empirical evidence to substantiate the performance and advantages of the proposed method.

\section{Supplementary Experiments}
\textbf{Experiments Details.} Table \ref{table_1} summarizes the data configurations of the DN-348 and DN-Wild benchmarks. DN-348 offers a balanced distribution of daytime and nighttime samples, whereas DN-Wild is substantially larger in scale and naturally exhibits a pronounced day–night imbalance. In addition, we report the computational characteristics of the proposed framework in Table \ref{table_2}. The overall training pipeline is outlined in Algorithm \ref{algorithm_1}.

\begin{table}[!h]
        \renewcommand{\arraystretch}{1.2}
	\belowrulesep=0pt
	\aboverulesep=0pt
	\centering
	\resizebox{\linewidth}{!}{
		\begin{tabular}{ccccc}
			\toprule[2pt] %change the first line to \toprule
			\multirow{1}{*}{\centering \textbf{Datasets}} & \multicolumn{1}{c}{Train(Day)} & \multicolumn{1}{c}{Train(Night)}  & \multicolumn{1}{c}{Test(Day)} & \multicolumn{1}{c}{Test(Night)}\\
			\cline{2-5}
                \hline
                DN-348     & 200/9962 & 200/10022 & 148/10121 & 148/3972 \\
                DN-Wild    & 1574/70981 & 1574/35384 & 712/14964 & 712/19568 \\
			\bottomrule[2pt] %change the third line to bottomrule
	\end{tabular}}
    \caption{Details of the dataset configuration, where */* denotes ID/Sample Size.}
    \label{table_1}
\end{table}

\begin{table}[!h]
	\centering
	\begin{tabular}{ccc}
            \toprule[2pt] %change the first line to \toprule
                \textbf{Modules} & FLOPs & Params\\
            \midrule %change the second line to midrule
            Image-Encoder     &14.70(G)& 85.65M            \\
                Text-Encoder    &1.94(G)         & 38.13M              \\
                Dynamic-bias Net  &143.36(K)           & 0.139M         \\
            \bottomrule[2pt] %change the third line to bottomrule
	\end{tabular}
	\caption{Computational complexity of the model components.}
    \label{table_2}
\end{table}

\begin{algorithm}[tb]
    \caption{The training procedure of proposed method.}
    \label{algorithm_1}
    
    \textbf{Input}: Unlabeled day-night dataset $D$, image encoder $E_{img}$, text encoder $E_{txt}$, Dynamic-Bias Net $f_{\theta}$\\
    \textbf{Stage-1}: 
    \begin{algorithmic}[1] %[1] enables line numbers
        \STATE Freeze $E_{img}$ and $E_{txt}$; initialize learnable prompt tokens $\{V_1, ..., V_M\}$
        \FOR{image $I_i$ in $D$ with epoch [0, epochs]}
        \STATE Extract visual features $f_i$ = $E_{img}(I_i)$ and compute instance-specific bias $f_\theta(f_i)$.
        \STATE Construct textual prompts according to Eq. 1 and obtain textual embeddings $t_i$ via $E_{txt}$.
        \STATE Compute bidirectional contrastive loss using Eq. 2-3.
        \ENDFOR 
        \STATE Save learned prompts and dynamic-bias net.
    \end{algorithmic}
    \textbf{Stage-2}: 
    \begin{algorithmic}[1] %[1] enables line numbers
        \STATE Perform DBSCAN clustering and initializing empty prototype memory banks $\{\phi_d,\phi_n\}$.
        \STATE Initialize domain-specific classifier parameters $\{\psi_d, \psi_n\}$ using the corresponding prototypes.
        \FOR{image $I$ in $D$ with epoch [0, iterations]}
            \STATE Update prototypes via momentum using Eq. 5.
            \STATE Calculate intra-domain loss: prototype contrastive loss ${\cal L}_{pcl}$ (Eq. 6) and pseudo-label classification loss ${\cal L}_{id}$ (Eq. 7-8).
            \STATE Recompute textual centroids within batch; Compute ${\cal L}_{i2tce}$ by Eq. 9.
            \STATE Construct cross-domain $k$-NN sets ${\cal R}(\phi_d), {\cal R}(\phi_n)$.
            \STATE Identify mutual nearest neighbors as positive pairs, with remaining neighbors treated as negatives.
            \STATE Compute cross-domain prototype matching loss ${\cal L}_{cpml}$ following Eq. 13.
            \STATE Update $E_{img}$ by optimizing the composite objective: ${\cal L} = {\cal L}_{id} + {\cal L}_{pcl} + {\cal L}_{i2tce} + {\cal L}_{cpml}$.
        \ENDFOR
        \STATE \textbf{return} $E_{img}$
    \end{algorithmic}
\end{algorithm}

\textbf{Generalization Ability of Individual Modules.} Table \ref{table_3} presents the cross-domain generalization performance of individual modules when transferring between DN-Wild and DN-348. The baseline already exhibits strong bidirectional transferability, suggesting that prototype-based representation learning inherently contributes to stable cross-domain generalization. Incorporating IPL yields further improvements in Rank-1 and mAP across all transfer settings by providing more reliable semantic alignment between visual instances and their dynamically adapted prompts. CPML alone also offers substantial gains by enforcing cross-domain prototype consistency, achieving more reliable cross-domain identity alignment. When combined, IPL and CPML deliver the most consistent performance enhancement, demonstrating their complementary roles in strengthening the model’s generalization ability under diverse day–night domain configurations.

\begin{table*}[!h]
	\belowrulesep=0pt
	\aboverulesep=0pt
	\centering
	\resizebox{\linewidth}{!}{
		\begin{tabular}{c|ccc|ccc|ccc|ccc|ccc}
			\toprule[2pt] %change the first line to \toprule
			\multirow{3}{*}{\centering \textbf{Index}} & \multicolumn{3}{c|}{\multirow{2}{*}{\centering \textbf{Components}}} & \multicolumn{6}{c|}{\textbf{DN-Wild$\rightarrow$DN-348}} & \multicolumn{6}{c}{\textbf{DN-348$\rightarrow$DN-Wild}}\\
			\cline{5-16} 
			 &&&&\multicolumn{3}{c|}{Day-to-Night } & \multicolumn{3}{c|}{Night-to-Day}& \multicolumn{3}{c|}{Day-to-Night} & \multicolumn{3}{c}{Night-to-Day}  \\
			\cline{2-16}
			& Baseline&IPL&CPML& Rank1 & Rank5 & mAP & Rank1 & Rank5 & mAP & Rank1 & Rank5 & mAP & Rank1 & Rank5 & mAP \\
			% \toprule[2pt] %change the second line to midrule
            \hline
            1& \Checkmark && &0.637 & 0.799 & 0.347 & 0.689 &0.829 & 0.352&0.462&0.930&0.234&0.423&0.859&0.242\\
            2& \Checkmark &\Checkmark&&0.655&\underline{0.810} & \underline{0.373}&\underline{0.723}&\underline{0.857}&0.362&0.470&\underline{0.935}&\textbf{0.244}&0.430&\textbf{0.884}&\textbf{0.256}\\
            3& \Checkmark &&\Checkmark&\underline{0.663}&0.804&0.365&0.689&0.839&\underline{0.363}&\underline{0.474}&\underline{0.935}&\underline{0.241}&\textbf{0.432}&0.878&0.252\\
            4& \Checkmark &\Checkmark&\Checkmark&\textbf{0.676}&\textbf{0.814}&\textbf{0.375}&\textbf{0.748}&\textbf{0.874}&\textbf{0.375}&\textbf{0.476}&\textbf{0.942}&0.240&\underline{0.431}&\underline{0.881}&\underline{0.254}\\
			\bottomrule[2pt] %change the third line to bottomrule
	\end{tabular}}
    \caption{Cross-domain generalization performance of individual modules, evaluating the impact of IPL and CPML when transferring between DN-Wild and DN-348.}
    \label{table_3}
\end{table*}

\textbf{Pseudo-label Accuracy.} As illustrated in Figure \ref{img_1}, we conduct a comprehensive comparison of clustering performance with several state-of-the-art methods on the DN-348 dataset. To ensure a fair and controlled evaluation, all approaches adopt the DBSCAN clustering algorithm with same hyperparameter settings; in particular, the maximum distance threshold eps is fixed to 0.7 for both domains. The specific clustering metrics are explained as follows:
\begin{itemize}
    \item ARI (Adjusted Rand Index) measures the similarity between two data partitionings.
    \item AMI (Adjusted Mutual Information) quantifies the agreement between clusters by normalizing shared information.
    \item FMI (Fowlkes-Mallows Index) evaluates the geometric mean of pairwise precision and recall.
    \item V-Measure computes the harmonic mean of completeness and homogeneity.
\end{itemize}
The proposed method (Ours) demonstrates superior performance across multiple clustering evaluation metrics, consistently outperforming recent approaches including RPNR, NULC, PCLHD, and PCA. This performance gain can be attributed to our synergistic integration of instance-aware prompt learning and cross-domain prototype matching, which effectively models the complex many-to-many correspondences between day and night clusters. The results validate that our framework establishes more robust identity associations under significant illumination shifts, bridging the domain gap more effectively than existing unsupervised methods.

\begin{figure}[!h]
    \centering
    \includegraphics[width=.5\textwidth]{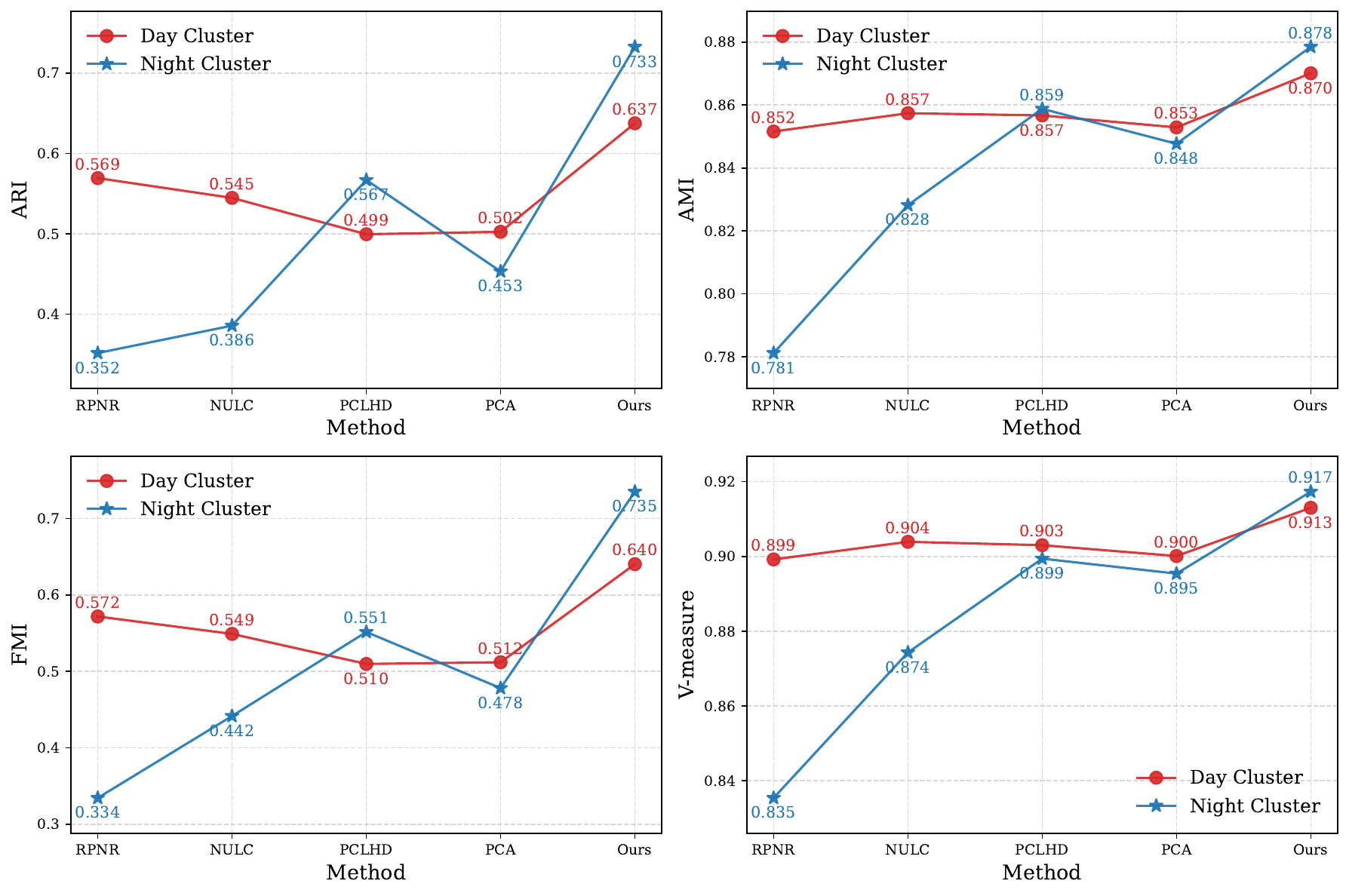}
    \caption{Performance comparison of clustering metrics against state-of-the-art methods on DN-348.}
    \label{img_1}
\end{figure}

\textbf{Visualization of Grad-CAM Activation Map.} Figure \ref{img_2} presents a comparison of Grad-CAM activation maps between the baseline and our method. The baseline model exhibits scattered and unstable attention, frequently responding to background clutter or illumination artifacts, particularly in nighttime scenes where glare and low-light noise dominate the visual field. In contrast, our method consistently concentrates on semantically meaningful and identity-relevant structures—such as headlights, grilles, and brand contours—across both day and night conditions. This stable focus indicates that the proposed framework alignment effectively suppress illumination bias and guide the network toward domain-invariant cues. 

\begin{figure}[!h]
    \centering
    \includegraphics[width=.5\textwidth]{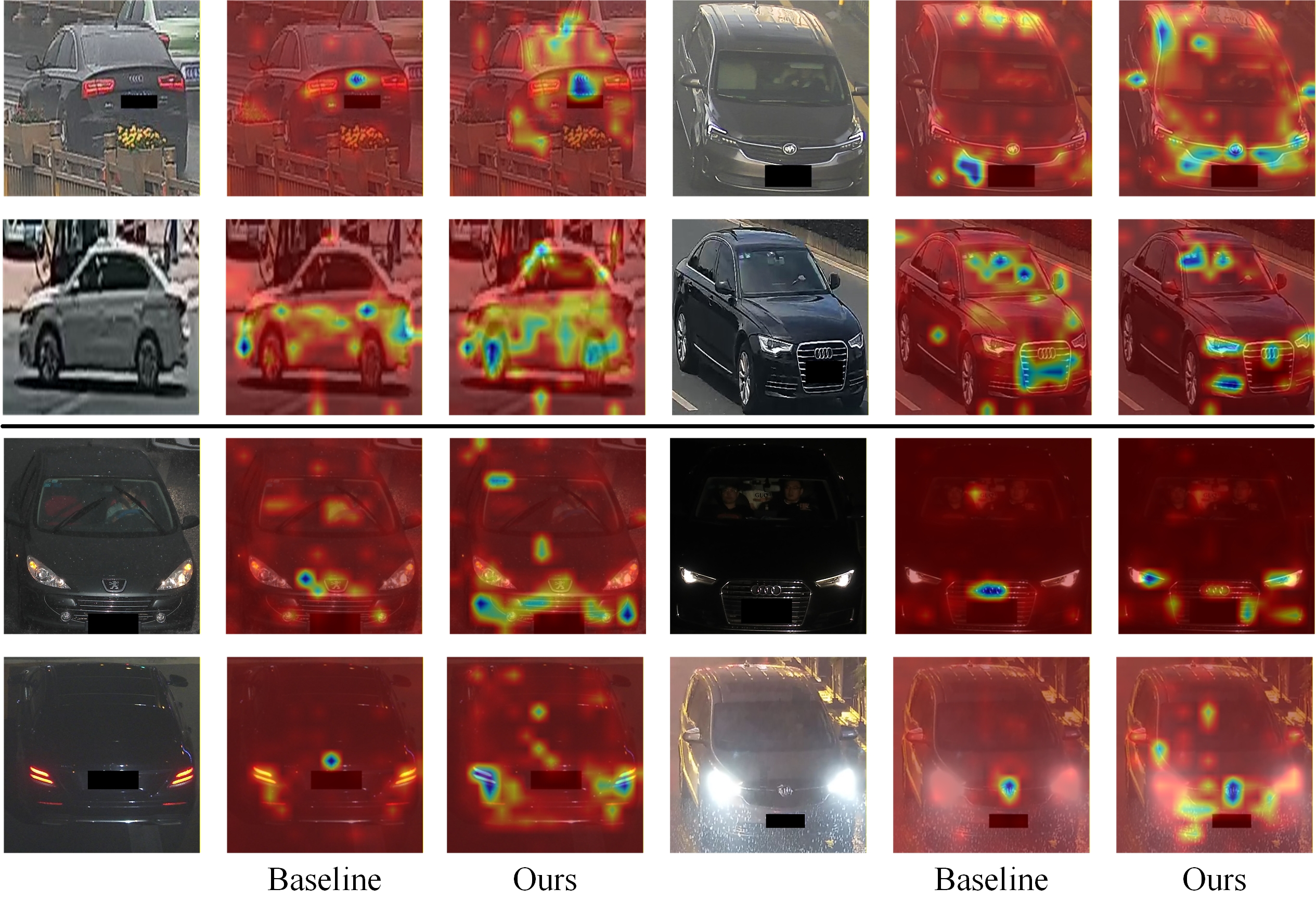}
    \caption{Comparative visualization of Grad-CAM activation maps for the baseline model and the proposed method.}
    \label{img_2}
\end{figure}

\textbf{Cross-domain Retrieval Ranklist.} Figure \ref{img_3} illustrates the top-5 cross-domain retrieval results for both day-to-night and night-to-day settings. Compared with the baseline, our method retrieves significantly more correct matches, especially under severe illumination changes where the baseline frequently confuses vehicles with similar colors or headlight patterns. The proposed approach consistently ranks the true identity at higher positions and maintains stable retrieval across varying viewpoints and lighting conditions. In challenging nighttime queries with glare or low-visibility regions, the baseline often fails to capture discriminative structure and produces visually similar but incorrect candidates, whereas our method preserves identity-relevant cues and suppresses illumination bias, leading to more accurate cross-domain association.
\begin{figure*}[!h]
    \centering
    \includegraphics[width=\textwidth]{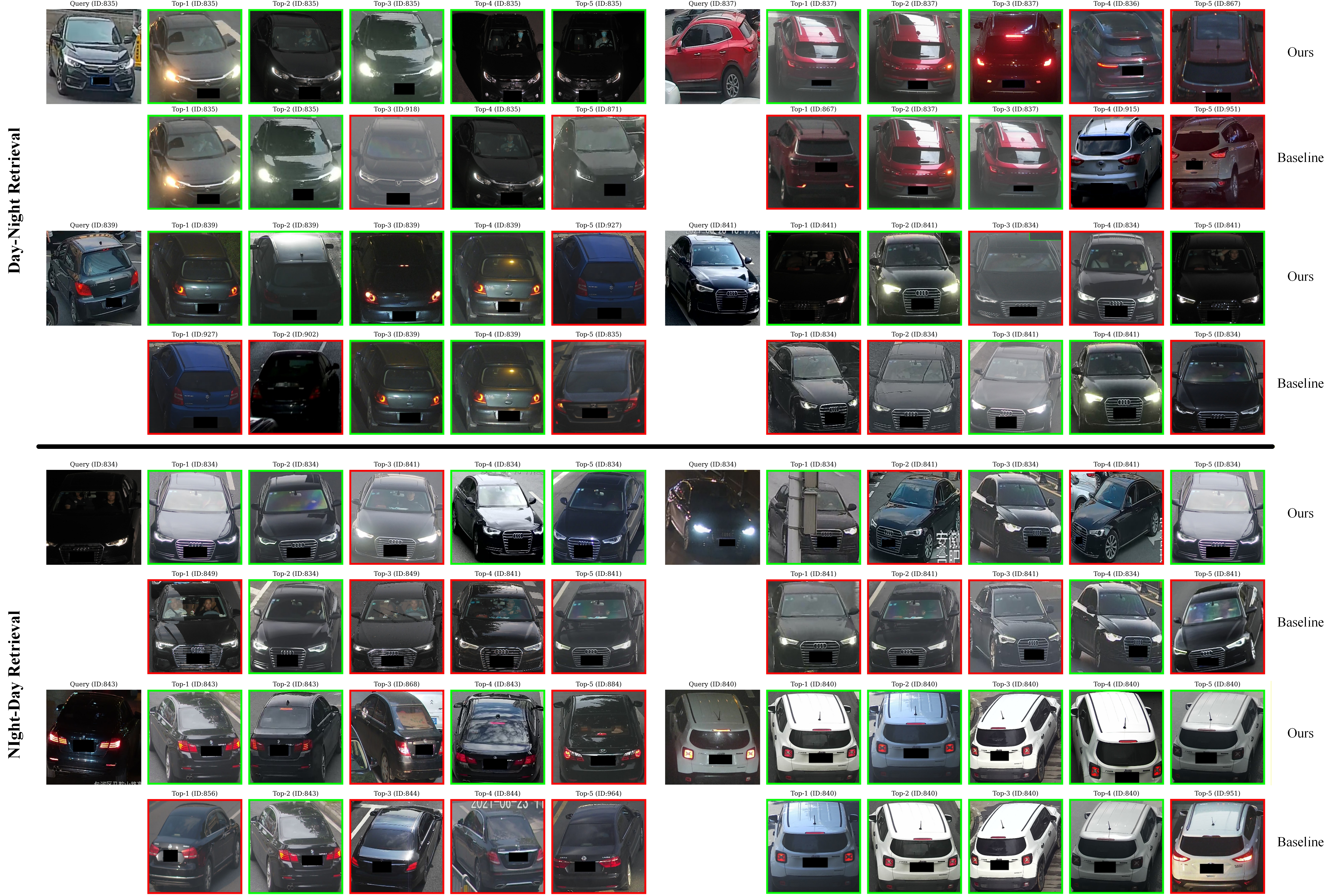}
    \caption{A comparison of the top-5 retrieval results under cross-domain day-night conditions is presented for the proposed method and the baseline. A green bounding box indicates a match with the ground truth identity, whereas a red box denotes a mismatch.}
    \label{img_3}
\end{figure*}

\section{Discussion}
Our work reveals several insights regarding the design choices and remaining challenges in unsupervised day–night vehicle re-identification. 
First, although the text encoder and prompt tokens are frozen in Stage-2, the textual representations remain dynamically updated through the instance-specific bias generated by the Dynamic-Bias Net. This mechanism effectively adapts the text features to the evolving visual representations, preventing semantic drift and preserving the benefits of Stage-1 alignment throughout the entire training pipeline.
Second, CPML constructs negative pairs from mutually non-matching cross-domain prototype neighbors, which naturally form the hardest negative samples in the latent space. These hard negatives enforce sharper prototype boundaries and encourage the model to refine cross-domain identity discrimination.

\end{document}